\documentclass[final,1p,times]{elsarticle}

\usepackage{graphicx} % Required for inserting images
\usepackage{bm}
\usepackage{multirow}
\usepackage{booktabs}
\usepackage{caption}
\usepackage{amssymb}
\usepackage{amsthm}  
  
\usepackage{mathrsfs}
\usepackage{mathtools}
\usepackage{diagbox}  % 放在导言区
\usepackage{titlesec}
\titlespacing*{\subsection}{0pt}{1.5ex plus .2ex minus .2ex}{1em}
\titleformat{\subsection}[runin]{\bfseries}{\thesubsection}{1em}{}
\titlespacing*{\subsection}{0pt}{\baselineskip}{\baselineskip}
\usepackage{amsmath}
\usepackage{float}
\usepackage{subcaption} 
\usepackage{algorithmic}
\usepackage{algorithm}
\usepackage{listings}
\usepackage{amsmath,amsfonts}
\usepackage{tikz}

\usepackage[labelfont=bf]{caption}
\usepackage{amssymb}
\usepackage{amsmath}
\usepackage[utf8]{inputenc}
\usepackage[colorlinks=true,    
linkcolor=blue,   
citecolor=green,
urlcolor=magenta, 
pdfborder={0 0 0} 
]{hyperref}

\begin{document}
\journal{Journal of Computational Physics}

\begin{frontmatter}



\title{Operator learning on domain boundary through combining fundamental solution-based artificial data and boundary integral techniques } %% Article title

%\author{Haochen Wu$^{1,2}$, Heng Wu$^{1,2}$, Benzhuo Lu$^{1,2,*}$ \\
%	$^1$ SKLMS, ICMSEC, NCMIS, Academy of Mathematics and Systems Science, Chinese Academy of Sciences, Beijing 100190, China \\
%	$^2$ School of Mathematical Sciences, University of Chinese Academy of Sciences, Beijing 100049, China \\
%	\texttt* {Correspondence: bzlu@lsec.cc.ac.cn}
%}

\author[inst1,inst2]{Haochen Wu}
\ead{wuhaochen@lsec.cc.ac.cn}
\author[inst1,inst2]{Heng Wu}
\ead{wuheng@amss.ac.cn}
\author[inst1,inst2]{Benzhuo Lu\corref{cor1}}

\cortext[cor1]{Corresponding author: bzlu@lsec.cc.ac.cn}

\address[inst1]{SKLMS, ICMSEC, NCMIS, Academy of Mathematics and Systems Science, Chinese Academy of Sciences, Beijing 100190, China}

\address[inst2]{School of Mathematical Sciences, University of Chinese Academy of Sciences, Beijing 100049, China}

%% Abstract

\begin{abstract}

	For linear partial differential equations (PDEs) with known fundamental solutions, this work introduces a novel operator learning framework. The key innovations are twofold:
	(1) Neural operators are trained exclusively using domain boundary data (i.e., solution values and normal derivatives), in contrast to most conventional methods that require full-domain sampling.
	(2) By integrating our previously developed Mathematical Artificial Data (MAD) method, which inherently enforces physical consistency, all training data are synthesized from the fundamental solutions of target problems. This enables a purely data-driven training pipeline that eliminates the need for external measurements or computational simulations.
	
	We term this approach the Mathematical Artificial Data–Boundary Neural Operator (MAD-BNO), which leverages MAD-generated Dirichlet–Neumann boundary data pairs to train a network for learning the boundary-to-boundary mapping. Once the neural operator is optimized, the interior solution at any point can be derived via boundary integral formulations, accommodating arbitrary Dirichlet/Neumann/mixed boundary conditions and source terms.
	
	MAD-BNO is validated on benchmark operator learning problems for  the 2D Laplace, Poisson, and Helmholtz equations. Experimental results show that it achieves accuracy comparable to or superior to other commonly used neural operator approaches, while reducing training time by a significant margin. The method is readily extendable to three-dimensional and complex geometries, as confirmed by the numerical examples we provide.
	﻿
\end{abstract}


%BO-net boundary operator network
%% Keywords
\end{frontmatter}



\section{Introduction}
\label{sec:introduction}
Partial differential equations (PDEs) are fundamental to modeling and analyzing a wide range of phenomena in science and engineering. They are widely used in fields such as fluid dynamics \cite{fluid11,fluid12,fluid13}, quantum mechanics \cite{quantum}, and materials science \cite{material}, providing critical insight into underlying physical principles. Classical mesh-based numerical methods---such as finite difference \cite{Randall-FDM}, finite element \cite{FEM}, and spectral methods \cite{spectral}---are well-established and effective. However, as problem scales expand---particularly in high-dimensional domains and geometrically complex systems---the computational and memory requirements of these methods escalate dramatically. Meanwhile, meshing/re-meshing processes for intricate geometries remain particularly burdensome, presenting substantial challenges in practical applications ranging from aircraft design \cite{plane,plane2} to biomolecular modeling
\cite{bio,bio2}.\par
With the advent of machine learning (ML) \cite{deeplearn} and its universal function approximation capability \cite{universe}, the learning of solutions to PDEs has been increasingly explored in recent years. While traditional methods remain the standard for solving well-posed PDEs with prescribed conditions, the accuracy and efficiency of many ML approaches remain debated, as highlighted in recent surveys and analyses \cite{WeakBaselinesNMI}. These studies emphasize potential issues such as weak baselines and reporting biases in the literature, which may lead to overoptimistic claims about performance. Such observations raise critical questions: Under what conditions can ML methods genuinely outperform state-of-the-art numerical solvers and which problem settings are most suitable for demonstrating these advantages? While this debate continues, one promising area is operator learning---learning mappings between infinite-dimensional input spaces (e.g., boundary/initial conditions or parameters) and solution fields. This approach is particularly valuable for tasks requiring efficient surrogate models, such as high-throughput batch computations, optimization, control, or uncertainty quantification.\par

Operator learning leverages ML to approximate complex high-dimensional relationships. Its potential spans nearly all practical domains due to its ability to handle diverse and large-scale applications. In recent years, operator learning has attracted significant attention, though real-world deployment remains limited. Methods in this field can be broadly categorized into model-driven (physics-informed) and data-driven approaches. Model-driven methods embed PDE residuals or conservation laws into the loss function (the model-driven treatment can be similar as in physics-informed neural networks (PINNs) \cite{PINN}). Data-driven approaches, by contrast, rely on supervised learning from precomputed numerical or experimental datasets to learn input-output mappings. Examples of purely data-driven operator learning frameworks include DeepONet \cite{deeponet}, Fourier Neural Operator (FNO) \cite{fno}, MiONet \cite{Mionet}, and MTL-DeepONet \cite{MTL}. Hybrid approaches, such as PI-DeepONet \cite{pIDEEP} and BI-GreenNet \cite{BIGREENNET}, combine elements of both paradigms. Purely model-driven is also possible, such as using the DeepONet structure while purely driven by the physics informed loss\cite{Cho2024PhysicsInformedDI}.\par 

Motivated by these developments and application requirements (e.g., in biomedical research), our work focuses on scalable operator learning for PDEs. In prior research, we introduced the Mathematical Artificial Data (MAD) paradigm \cite{MAD}, which transforms model-driven approaches into data-driven ones by efficiently generating synthetic, precise physics-embedded training datasets through explicit expressions. This approach accelerates training and avoids reliance on costly numerical simulations or experiments. However, like most existing operator learning models, MAD relies on dense interior sampling and domain-wide mean-square-error (MSE) minimization.\par

This work targets linear PDEs with fundamental solutions, a class encompassing many elliptic equations under certain conditions,  which are central to modeling steady-state phenomena, wave propagation, and potential theory. For such PDEs, we propose a Mathematical Artificial Data–Boundary Neural Operator (MAD-BNO). Leveraging classical boundary integral theory, MAD-BNO reformulates the PDE into a boundary integral term and a deterministic source integral term (see Methodology). Consequently, both known and unknown information are confined to the boundary, reducing the learning task to a boundary-aware operator. Concurrently, this work employs a methodology analogous to MAD's artificial data generation framework, generating boundary-aligned training data through the utilization of PDE fundamental solutions\cite{cheng2025introduction}. This eliminates the need for interior sampling, enabling reconstruction of the full domain solution from boundary data alone. The outcome is an operator learning framework characterized by certain theoretical basis, superior computational efficiency, mesh-free implementation, and exceptional geometric adaptability---owing to simplified boundary processing compared to volume domain processing.
\par

The remainder of this paper is organized as follows. Section~\ref{sec:methodology} details the methodology of
MAD-BNO and the data generations in current and compared methods; Section~\ref{sec:numerical_results} presents numerical results; and Section~\ref{sec:discussion} offers discussion and future outlook.

\section{Methodology}
\label{sec:methodology}
\subsection{Integral Representation}
This work focuses on linear PDEs that admit fundamental solutions. Such solutions form the basis for integral representations in potential theory. The equation with a general boundary conditions (BCs) is written as follows:
\begin{equation}
    \begin{cases}
        \mathcal{L} u(x)=f(x),x\in \Omega,\\
        u(x)\big|_{\partial\Omega_{D}}=g_D(x),\\
        \frac{\partial u}{\partial n}(x)\big|_{\partial \Omega_{N}}=h_N(x),\label{origineq}
    \end{cases}
\end{equation}
where $\mathcal{L}$ is a linear differential operator and $\Omega$ is a given domain, and where $\partial\Omega_{D} \cup \partial\Omega_{N} = \partial \Omega$ denotes the full boundary, with Dirichlet and Neumann conditions prescribed on $\partial\Omega_D$ and $\partial\Omega_N$, respectively. Dirichlet and Neumann BCs are special cases of mixed BCs, and the methods are the same in our work. In this study,
we aim to learn the solution operator mapping:
\begin{equation}
\mathcal{G}:(f, g_D, h_N) \mapsto u,
\end{equation}
This work is motivated by the observation that canonical elliptic PDEs (Laplace, Poisson and Helmholtz as illustrated in this work) can be reformulated as boundary integral equations comprising two components:
\begin{itemize}
	\item A volume integral \( \int_{\Omega} G(x,y)f(y) dy \), encoding source term contributions, which is analytically tractable via numerical quadrature.
	\item A boundary integral dependent solely on BCs \( u|_{\partial \Omega},\ \frac{\partial u}{\partial n}|_{\partial \Omega} \).
\end{itemize}

The integral representation reads\cite{ELIpde}:
\begin{equation}
	u(x) = \int_{\partial \Omega} \left( u(y) \frac{\partial G(x, y)}{\partial n_y} - G(x, y) \frac{\partial u(y)}{\partial n_y} \right) \mathrm{d}s(y)+ \int_{\Omega} G(x,y)f(y) dy
	\label{eq:integral1}
\end{equation}
where \( G \) denotes the fundamental solution of the elliptic operator \( \mathcal{L} \).
Owing to the well-posedness and uniqueness of the mixed boundary value problems(BVPs), Eq.~\ref{origineq} admits a unique solution under suitable regularity assumptions. Therefore, the task reduces to accurately reconstructing the unknown  boundary data so as to complete the specification on $\partial \Omega$. Once the full boundary information is recovered, the solution in the entire domain $\Omega$ can be uniquely determined via the boundary integral representation. Therefore, we actually aim to learn the mapping
\begin{equation}
	\mathcal{N}_\theta :\left(u\big|_{\partial\Omega_{D}},\frac{\partial u}{\partial n}\big|_{\partial\Omega_{N}}\right) \mapsto \left(\frac{\partial u}{\partial n}\big|_{\partial\Omega_{D}}, u\big|_{\partial\Omega_{N}}\right).
\end{equation}
Here, \( \mathcal{N}_\theta \) is a neural operator parameterized by \( \theta \), trained to map the available Dirichlet and Neumann data on complementary boundary segments to the unknown boundary data in a mixed boundary value problem.

\par For notational simplicity, we denote \( \frac{\partial u}{\partial n}\big|_{\partial\Omega_D} \) as \( h_D \) and \( u\big|_{\partial\Omega_N} \) as \( g_N \). In the sequel, we use the simplified notation
\begin{equation}
	\mathcal{N}_\theta : (g_D, h_N) \mapsto (h_D, g_N).
\end{equation}
We focus on three representative linear elliptic PDEs for demonstration: the Laplace equation (Eq.~\ref{eq:laplace}) with general mixed boundary conditions, the source-free Helmholtz equation (Eq.~\ref{eq:helmholtz}), and the Poisson equation (Eq.~\ref{eq:poisson}) with Dirichlet boundary conditions (similar treatments to mixed BCs). These problems have wide applications in fields such as electromagnetism~\cite{elect} and acoustics~\cite{acoustic}. For the Helmholtz and Poisson equations, the boundary integral formulation can also handle general mixed BCs, which we omit here for brevity.
These boundary value problems are given as follows.  

The Laplace boundary value problem is  
\begin{equation}\label{eq:laplace}
	\left\{
	\begin{aligned}
		&\Delta u(x) = 0, \quad x \in \Omega, \\
		&u(x)|_{\partial\Omega_D} = g_D(x), \\
		&\frac{\partial u}{\partial n}(x) \big|_{\partial \Omega_N} = h_N(x),
	\end{aligned}
	\right.
\end{equation}
and a two-dimensional fundamental solution of the Laplace operator:
\begin{equation}\label{eq:green_laplace}
	G_{\text{L}}(x,y) = -\frac{1}{2\pi} \ln |x - y|.
\end{equation}

The Helmholtz boundary value problem is  
\begin{equation}\label{eq:helmholtz}
	\left\{
	\begin{aligned}
		&\Delta u(x) + k^{2} u(x) = 0, \quad x \in \Omega, \\
		&u(x)|_{\partial \Omega} = g(x),
	\end{aligned}
	\right.
\end{equation}
with the corresponding fundamental solution  of the Helmholtz operator:
\begin{equation}\label{eq:green_helmholtz}
	G_{\text{H}}(x,y) = \frac{i}{4} H_0^{(1)}(k |x - y|), \quad k>0,
\end{equation}
where $H_0^{(1)}$ denotes the Hankel function of the first kind of order zero, i.e.,
\[
H_0^{(1)}(z) = J_0(z) + i Y_0(z),
\]
with $J_0$ and $Y_0$ being the Bessel functions of the first and second kind of order zero, respectively.
\par
The Poisson  boundary value problems is expressed as  
\begin{equation}\label{eq:poisson}
	\left\{
	\begin{aligned}
		&\Delta u(x) = f(x), \quad x \in \Omega, \\
		&u(x)|_{\partial \Omega} = g(x),
	\end{aligned}
	\right.
\end{equation}
The fundamental solution for the Poisson equation is the same as that for the Laplace equation . We further decompose the solution of Eq.~\ref{eq:poisson} into two parts:
\begin{equation}\label{2}
	u(x) = u_f(x) + u_g(x),
\end{equation}
\begin{equation}\label{eq:integral2}
	u_f(x) = \int_{\Omega} G(x, y) f(y) \, \mathrm{d}y,
\end{equation} 
\begin{equation}\label{eq:laplacetrans}
	\left\{
	\begin{aligned}
		&\Delta u_g(x) = 0, \quad x \in \Omega, \\
		&u_g(x)|_{\partial \Omega} = g(x) - u_f(x)|_{\partial \Omega}.
	\end{aligned}
	\right.
\end{equation}

where \( u_f \) denotes a known part from source contribution $f(x)$ and the unknown part $u_g$ satisfies a Laplace  boundary value problem (Eq.~\ref{eq:laplacetrans}).
Through this decomposition, the Poisson problem is transformed into a Laplace-type boundary value problem, which can subsequently be solved using the neural operator framework introduced earlier. Extending this methodology, analogous decomposition techniques may be applied to a broader class of PDEs with source terms, effectively reformulating them as source-free problems.
                                                                                                                                                                                                                                           
\subsection{ Boundary-aware Neural Operator Learning via Fundamental Solutions}
%Accordingly, we focus on learning the complementary boundary data
%\[
%\mathbf{b}_{\text{unknown}} = (h_{D},g_{N})
%\]
%
%
% These missing components are predicted through a neural operator \( \mathcal{N}_\theta \) parameterized by \( \theta \):
%\[
%\mathbf{b}_{\text{unknown}} = \mathcal{N}_\theta(\mathbf{b}_{\text{known}}).
%\]
%
%This boundary-to-boundary learning framework significantly reduces the traditional domain-wise learning space. The complete workflow comprises three stages:
%\begin{enumerate}
%	\item \textbf{Data Preprocessing:} Extract boundary traces from domain solutions.
%	\item \textbf{Operator Learning:} Train \( \mathcal{N}_\theta \) on boundary pairs \( (\mathbf{b}_{\text{known}}, \mathbf{b}_{\text{unknown}}) \).
%	\item \textbf{Solution Reconstruction:} Compute \( u(x) \) via the boundary integral using the combined boundary data \( \mathbf{b}_{\text{known}} \cup \mathbf{b}_{\text{unknown}} \).
%\end{enumerate}
﻿
As noted in the Introduction, our previous work on the MAD framework demonstrated that using analytical solutions to construct training datasets substantially reduces the cost of generating high-quality, diverse data for operator learning. Building on this idea, we further observe that for source-free problems such as the Laplace and Helmholtz boundary value problems, the fundamental solutions themselves are exact  solutions that naturally satisfy the governing PDEs. For source-driven problems such as the Poisson boundary value problems, a decomposition strategy allows the original problem to be reformulated as the superposition of a particular solution (accounting for the source term) and a homogeneous solution. In particular, the source-induced response can be accurately computed via a volume integral involving the fundamental solution, while the remaining component satisfies the corresponding homogeneous PDE.
﻿
Motivated by this insight, we construct our entire training dataset using fundamental solutions and their linear combinations, enabling the generation of exact Dirichlet–Neumann boundary data pairs for supervised learning.
This approach is not only analytically tractable but also theoretically justified: the linear span of such fundamental solutions has been shown to be dense in the relevant solution space, particularly in 
$L^{2}(\partial \Omega)$ \cite{dense}.
This guarantees the expressiveness of our dataset and the potential of the learned model to generalize across a wide class of boundary conditions.

%\subsection{Model Training and Loss Definitions}
After introducing the boundary integral formulation and the role of fundamental solutions, we now turn to the learning objectives that govern model training.  
Broadly speaking, existing PDE learning frameworks can be categorized as either model-driven or data-driven, based on how they define their loss functions.
To contextualize our method within this context, we briefly review the typical loss formulations used in both approaches.
\begin{equation}
	\mathcal{L}_{\text{data}}
	= \frac{1}{N} \sum_{i=1}^{N} \bigl\lVert u^{\mathrm{NN}}(x_{i}) - u^{\mathrm{data}}(x_{i}) \bigr\rVert^{2},
\end{equation}
where $u^{\mathrm{NN}}$ and $u^{\mathrm{data}}$ denote predicted and reference solutions at points $x_i$. By contrast, model-driven methods typically incorporate the PDE residual:
\begin{equation}
	\mathcal{L}_{\mathrm{model}}
	= \frac{1}{M} \sum_{j=1}^{M} \bigl\lVert \mathcal{L} u^{\mathrm{NN}}(x_{j}) - f(x_{j}) \bigr\rVert^{2},
\end{equation}
Within the framework of data-driven methods, our approach constructs the loss function primarily from a data fidelity term measuring the difference between predicted and true values. The  loss function under mixed boundary conditions is defined as:
\begin{equation}
	\begin{aligned}
		&\mathcal{L}_{\mathrm{data}}=\frac{1}{N_{1}+N_{2}}\bigg(\lambda_{1}\sum_{i=1}^{N_1}\Vert \mathcal{N}_\theta\big(g_D(x_{i})\big)-h_{D}(x_{i})\Vert_{2}^{2}\\&+
		\lambda_{2}\sum_{j=1}^{N_2}\Vert \mathcal{N}_\theta\big(h_N(x_{j})\big)-g_{N}(x_{j})\Vert_{2}^{2}\bigg),~~~~x_{i}\in \partial \Omega_{D}~,~x_{j}\in \partial \Omega_{N}.
	\end{aligned}
\end{equation}
For mixed boundary conditions, the weights $\lambda_1$ and $\lambda_2$ in the loss function serve to balance the contributions from Dirichlet and Neumann boundary data. Since these two types of data generally differ in physical units and magnitude, $\lambda_1$ and $\lambda_2$ can be adjusted in practice to ensure that the training process is comparably sensitive to both boundary types. In this work, for simplicity and methodological uniformity, we set $\lambda_{1}=1$,$\lambda_{2}=1$.
%The proposed network is designed to accept either Dirichlet or Neumann boundary data as input and infer the corresponding complementary boundary condition on the domain boundary. \par
\subsection{Dataset Generation for MAD-BNO}
\label{sec:data_generation} 
The given boundary conditions \( g_D \), \( h_N \), along with their accurate complementary counterparts \( h_D \), \( g_N \), play a crucial role in determining the accuracy and generalization of the trained operator.
For pure Dirichlet problems, we set \(\partial\Omega_{D} = \partial\Omega\). 
Therefore, constructing sufficiently varied and representative boundary samples is crucial. 
Below, we describe our strategies for generating training data for Laplace(similar to Poisson)  and Helmholtz boundary value problems.

\subsubsection{Laplace and Poisson Equations}

To generate diverse Dirichlet and Neumann boundary data for the Laplace equation, we define the boundary input as a linear combination of analytically constructed harmonic functions:
\begin{equation}\label{eq:harmonic-combo}
	u(x,y)
	\;=\;
	\sum_{i=1}^{n} c_i\,f_i(x,y),
	\quad
	\sum_{i=1}^n c_i = 1,
\end{equation}
Each function \( f_i(x, y) \) is chosen as a fundamental solution of the two-dimensional Laplace equation, specifically a logarithmic kernel of the form  
\[
f_i(x, y) = \log\left( (x - x_i)^2 + (y - y_i)^2 \right),
\]  
where \( (x_i, y_i) \) denotes the source point associated with the \(i\)-th kernel. To ensure the well-posedness of the solution and avoid singularities within the computational domain \(\Omega\), the source points are uniformly sampled in a bounding box \([-7, 7] \times [-7, 7]\) (strictly outside $\Omega$) and constrained to lie at least \( h = 10^{-3} \) away from the boundary \( \partial\Omega \).

The combination weights \( c_i \) are randomly sampled and normalized to satisfy the constraint \( \sum_{i=1}^n c_i = 1 \). This normalization prevents excessive amplitudes in the input functions and promotes numerical stability during training.
Specifically, the coefficients are generated sequentially using a simple rejection-free procedure: the first weight \( c_1 \) is sampled uniformly from the interval \( [0, 1] \), and each subsequent weight \( c_i \) is sampled uniformly from \( \left[0, 1 - \sum_{j=1}^{i-1} c_j \right] \) for \( i = 2, \dots, n \). The final weight ensures exact normalization:
\[
c_n = 1 - \sum_{j=1}^{n-1} c_j,~n=3.
\]
This procedure ensures that the coefficients form a convex combination and lie on the probability simplex. The corresponding normal derivatives \( \frac{\partial f_i }{\partial n} \) on the boundary \( \partial\Omega \) are computed analytically, enabling the generation of high-fidelity Dirichlet–Neumann boundary data pairs for supervised learning.

The normal derivative is obtained by projecting the gradient of \( f_i \) onto the outward unit normal vector \( \mathbf{n} = (n_x, n_y) \) at the boundary:
\begin{equation}
\frac{\partial f_i}{\partial n} = \nabla f_i \cdot \mathbf{n} = \frac{\partial f_i}{\partial x} n_x + \frac{\partial f_i}{\partial y} n_y.\label{normal}
\end{equation}
For the Poisson boundary value problem, as shown in methodology (Eqs.~\ref{eq:integral2}–\ref{eq:laplacetrans}), the problem is reduced to a Laplace boundary value problem for the harmonic component \(u_g\). Consequently, our model still only requires Laplace‐equation boundary data for training.

\subsubsection{ Helmholtz Equation}

For the Helmholtz operator, we use its fundamental solution, which can be expressed in terms of the zeroth-order Bessel functions of the first and second kinds ($J_0$ and $Y_0$). These functions are defined as:

\begin{equation}
	J_{0}(r)=\sum\limits_{m=0}^{\infty}\frac{(-1)^{m}}{(m!)^{2}}\bigl(\frac{r}{2}\bigr)^{2m},
\end{equation}
\begin{equation}
	Y_{0}(r)=\frac{2}{\pi}\Biggl[J_{0}(r)\ln\Bigl(\frac{r}{2}\Bigr)
	-\sum\limits_{m=1}^{\infty}\frac{(-1)^{m}}{(m!)^{2}}\Bigl(\frac{r}{2}\Bigr)^{2m}\bigl(\psi(m)+\psi(m+1)\bigr)\Biggr],
\end{equation}
\begin{equation}
	\psi(m) = \frac{\mathrm{d}}{\mathrm{d}m} \ln \Gamma(m) = \frac{\Gamma'(m)}{\Gamma(m)}, \quad m \in \mathbb{C} \setminus \{0, -1, -2, \ldots\}
\end{equation}
where \(\psi(m)\) is the digamma function, \(r = \lvert (x, y)-(x_{i}, y_{i})\rvert\) represents the distance between the current point \((x, y)\) and the external point \((x_{i}, y_{i})\). The functions \(J_{0}(kr)\) and \(Y_{0}(kr)\), which denote the real and imaginary parts of the Hankel function of the first kind, are used as the fundamental solutions for the 2D Helmholtz equation in our framework.

To obtain accurate ground truth labels, we utilize the derivative properties of the zeroth-order Bessel functions. Specifically, for the fundamental solutions of the 2D Helmholtz equations, the derivatives are given by\cite{partialde}:
\begin{equation}
	\begin{aligned}
		\frac{\partial}{\partial x} J_0(kr) = -k \frac{x}{r} J_1(kr)~,~ &
		\frac{\partial}{\partial y} J_0(kr) = -k \frac{y}{r} J_1(kr),\\
		\frac{\partial}{\partial x} Y_0(kr) = -k \frac{x}{r} Y_1(kr)~,~ &
		\frac{\partial}{\partial y} Y_0(kr) = -k \frac{y}{r} Y_1(kr),
	\end{aligned}
\end{equation}
Here, \(J_1\) and \(Y_1\) denote the first-order Bessel functions of the first and second kinds, respectively. Specifically, \(J_1(kr)\) and \(Y_1(kr)\) are the derivatives of \(J_0(kr)\) and \(Y_0(kr)\),  and describe the rate of change of the oscillatory behavior of the fundamental solutions. The normal derivatives are computed analogously to the Laplace case (Eq.~\ref{normal}).
These expressions are essential for computing the normal derivatives along the boundary, enabling the generation of precise Neumann boundary condition labels.
\subsection{Experimental Setup for MAD-BNO}
\label{sec:data_generation2} 
In our benchmark evaluations, we systematically compare MAD-BNO with PI-DeepONet and our previous MAD framework for the Laplace, Poisson, and source-free Helmholtz boundary value problems. In our previous MAD framework\cite{MAD}, artificially synthetic datasets and a neural operator architecture analogous to DeepONet\cite{deeponet} were employed for training. Given this methodological lineage, a formal method name
of that could be MAD-DeepONet. For brevity in comparative analysis, we uniformly refer to that approach as MAD in subsequent discussions. Computational efficiency is quantified by wall-clock time per \(50{,}000\) training epochs, and accuracy by relative \(L^2\) error (see Section~\ref{sec:numerical_results} for details).

We construct a boundary condition dataset with \(10^4\) analytical functions, each sampled at 400 collocation points uniformly distributed along the boundary \(\Gamma = \partial\Omega\). Training employs the Adam optimizer \cite{Adam} with a learning rate of \(10^{-4}\) and batch size \(10^3\). A best-model checkpoint strategy saves parameters whenever the training loss reaches a new minimum.\par
To reduce the functional approximation space in Dirichlet boundary value problems, we first normalize the Dirichlet boundary data by subtracting its first entry from all components, thereby setting the first element to zero. Next, to control the magnitude of both Dirichlet and Neumann data and avoid numerical instability caused by large Neumann values, we divide both vectors by the maximum absolute value of the Neumann vector. This two-step normalization keeps all inputs and outputs within a stable numerical range. Since our network is fully linear without bias terms, these transformations preserve the correctness of the underlying Dirichlet-to-Neumann mapping: both the translation and rescaling can be equivalently represented within the same architecture.
\par All experiments were implemented in Python 3.12.5 using PyTorch 2.4.1+cu124, running on the ORISE supercomputing system.
\subsection{Network Architecture of MAD-BNO}
For boundary value problems with Dirichlet, Neumann, or mixed boundary conditions, we exploit the linearity of the corresponding boundary-to-boundary operator and approximate it using a bias-free, fully connected linear network. The input is a vector of boundary values sampled at \(n=400\) uniformly distributed collocation points, and the output is the corresponding vector of boundary responses (normal derivatives for DtN, boundary values for NtD, etc.). Based on the relative \(L^2\) errors reported in Table~\ref{tab:depth_width_loss}, we adopt a \([400, 400]\) configuration---a single linear layer mapping 400 inputs to 400 outputs---which provides an optimal trade-off between accuracy and architectural simplicity.\par
﻿
The resulting \(400 \times 400\) weight matrix can be regarded as a discrete representation of the underlying continuous boundary operator. Importantly, this matrix is entirely determined by the boundary geometry and the differential operator defining the PDE, and it is independent of the specific boundary data. Consequently, for a given domain and underlying PDE, the network realizes a deterministic linear operator that maps boundary data to the corresponding boundary responses, encompassing Dirichlet-to-Neumann, Neumann-to-Dirichlet, or mixed-type boundary conditions.
﻿

\subsection{Data Sampling in PI-DeepONet Methods}
\label{sec:data_generation3} 
To compare with our method, we considered PI-DeepONet \cite{pIDEEP} applied to the Dirichlet problems for the Laplace, Poisson, and source-free Helmholtz equations. For a fair comparison, we used \(10^4\) randomly generated functions as the training set, matching the dataset size used in our approach. Training datasets were constructed by independently sampling boundary conditions \(g\) and source terms \(f\), ensuring diversity and representativeness for robust model training. The network architecture used for PI-DeepONet follows the configuration described in \cite{MAD}.

\subsubsection{Boundary Condition Sampling}
We constructed the boundary conditions dataset by sampling raw values \(g_{\mathrm{raw}}\) from a Gaussian random field (GRF) \cite{GRF} with zero mean:
\[
g_{\mathrm{raw}}(x) \sim \mathcal{G}\bigl(0,\kappa_{l}(x,x')\bigr),\quad
\kappa_{l}(x,x') = \exp\!\Bigl(-\frac{\lvert x - x' \rvert^2}{2l^{2}}\Bigr),\quad
l = 0.1.
\]
To enforce periodicity at the boundary endpoints \(x_1\) and \(x_N\), we compute
\[
\delta = g_{\mathrm{raw}}(x_{N}) - g_{\mathrm{raw}}(x_{1}),
\quad
g(x_i) = g_{\mathrm{raw}}(x_i) - \delta\,\frac{i-1}{N-1},\quad i=1,\dots,N,
\]
so that \(g(x_1)=g(x_N)\), while preserving the GRF variability. Here, $g_{\mathrm{raw}}$ represents the independently sampled raw boundary values, and $g$ denotes the adjusted boundary values used for training after enforcing periodicity.

\subsubsection{Source Term Sampling}
First, we generate raw source samples \(f_{\mathrm{raw}}(x_j)\sim\mathcal{N}(0,1)\) independently at each grid point \(x_j\). To introduce spatial smoothness, we convolve these samples with a Gaussian kernel over the discretized domain as follow:
\[
f_{\mathrm{smoothed}}(x_i)
= \sum_{j=1}^{N_x N_y}
f_{\mathrm{raw}}(x_j)\,
\frac{1}{2\pi\sigma^2}
\exp\!\Bigl(-\frac{\lvert x_i - x_j\rvert^2}{2\sigma^2}\Bigr)\,\Delta A,
\quad \sigma = 5,
\]
where $\Delta A$ is the area of one grid cell, $f_{\mathrm{raw}}$ denotes the independently sampled raw source values at each grid point, and $f_{\mathrm{smoothed}}$ represents the spatially smoothed version used for training; this ensures continuity and smoothness in the training data.

\subsubsection{PI-DeepONet Loss Function}
We employed a DeepONet \(u^{\mathrm{NN}}\) to approximate the PDE solution, with the residual \(\mathcal{L} u^{\mathrm{NN}}\) computed using automatic differentiation. The loss function combines the PDE residual and boundary conditions:
\begin{equation}
	\mathcal{L_{PI}}
	= \sum_{i=1}^{N}\bigl\lVert \mathcal{L} u^{\mathrm{NN}}(x_i,y_i)-f(x_i,y_i)\bigr\rVert_{2}^{2}
	+ \lambda \sum_{j=1}^{M}\bigl\lVert u^{\mathrm{NN}}(x_j,y_j)-g(x_j,y_j)\bigr\rVert_{2}^{2},
\end{equation}
where \(\lambda\) is a tunable hyperparameter (set to \(0.1\) in our experiments); 
\(\{(x_i,y_i)\}_{i=1}^N\) are the interior collocation points in \(\Omega\); 
\(\{(x_j,y_j)\}_{j=1}^M\) are the boundary collocation points on \(\partial\Omega\); 
and \(g\) represents the given Dirichlet boundary values. 
In our experiments, the collocation points are sampled from a uniform $51\times51$ grid. 
This purely physics-driven formulation does not require labeled solution data.
\subsection{Data Sampling in MAD Methods}
 In the square domain, the collocation points in MAD coincide with those used in PI-DeepONet. The specific sampling procedure of MAD is detailed in our previous work \cite{MAD}.

\subsection{Singular Integral Volume and boundary integral}
\label{singularintegral}
Once the training of the above boundary operators
is completed, the solution at any position within the
domain can be obtained through boundary integration and volume integration (if there is a source term) (Eq.~\ref{eq:integral1}). We have implemented Python  code for 2D volume and boundary numerical integrals.

\subsubsection{Volume Integration Involving Singular Kernel}

The key distinction between the Poisson and Laplace equations lies in the additional singular integral term in Eq.~\ref{eq:integral2}. 
To evaluate this integral, we generated a uniform triangular mesh over the domain using Gmsh (Version 4.13.1)\cite{Gmsh}, aiming for an element size of $h=0.01$. 
For example, for a square domain, this resulted in a total of 23,258 triangular elements. 
Near each singular point, a small circular neighborhood (radius $r_0 = 0.01$) is treated with a polar coordinate transformation to analytically handle the logarithmic singularity, 
while the rest of the domain is integrated using standard 2D Gaussian quadrature.
﻿

To accurately evaluate the singular double integral with a logarithmic kernel, we partition the integration domain into two regions. Within a circular neighborhood (radius \( r_0 = 0.01 \)) centered at the singular point, polar coordinate transformation is used to isolate and analytically eliminate the logarithmic singularity. Outside this neighborhood, integrals over each triangle are computed using 2D seven-point Gaussian quadrature \cite{Gaussweight}, ensuring high accuracy and fast convergence.

\begin{figure}[h]
	\centering
\begin{tikzpicture}[scale=2]
	% 统一边线颜色和粗细
	\tikzset{triangle_style/.style={thick, black}} 
	
	% 左边三角形 T
	\coordinate (v1) at (0.2,0.1);
	\coordinate (v2) at (0.9,0.2);
	\coordinate (v3) at (0.4,0.7);
	
	\draw[triangle_style] (v1) -- (v2) -- (v3) -- cycle;
	\node at (v1) [below left] {$\mathbf{v}_1$};
	\node at (v2) [below right] {$\mathbf{v}_2$};
	\node at (v3) [above left] {$\mathbf{v}_3$};
	\node at (0.55,0.45) {$T$};
	
	% 右边单位参考三角形 \hat{T}
	\begin{scope}[xshift=3.5cm]
		\draw[triangle_style] (0,0) -- (1,0) -- (0,1) -- cycle;
		\node at (0,0) [below left] {$(0,0)$};
		\node at (1,0) [below right] {$(1,0)$};
		\node at (0,1) [above left] {$(0,1)$};
		\node at (0.35,0.15) {$\hat{T}$};
	\end{scope}
	
	% 箭头
\draw[->, thick] (1.3,0.4) -- (3,0.4) node[midway, above] {Affine map}; 
\end{tikzpicture}
	\caption{Mapping a general triangle \(T\) with vertices \(\mathbf{v}_1, \mathbf{v}_2, \mathbf{v}_3\) to the reference triangle \(\hat{T}\).}
	\label{fig:ref_triangle}
\end{figure}
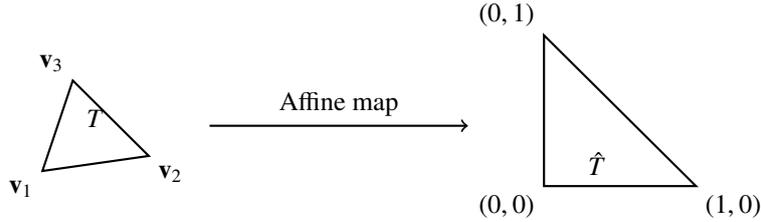

To perform Gaussian quadrature over a general triangle \( T \) with vertices \( \mathbf{v}_1 = (x_1, y_1) \), \( \mathbf{v}_2 = (x_2, y_2) \), and \( \mathbf{v}_3 = (x_3, y_3) \), we use the affine mapping between \( T \) and the reference triangle \( \hat{T} \) as shown in Figure~\ref{fig:ref_triangle}. The transformation is given by:

\[
\mathbf{x}(\xi, \eta) = \mathbf{v}_1 + (\mathbf{v}_2 - \mathbf{v}_1)\xi + (\mathbf{v}_3 - \mathbf{v}_1)\eta,
\]
where \( (\xi, \eta) \in \hat{T} \) with \( \xi \geq 0 \), \( \eta \geq 0 \), and \( \xi + \eta \leq 1 \). Here, $\mathbf{x} = (x, y)$ represents the 2D coordinates of a point within the triangle $T$, and $\mathbf{v}_i = (x_i, y_i)$ are the vertices of the triangle. The boldface indicates that these are vector quantities rather than scalar values.

Thus, the integral of a function \( f \) over the triangle \( T \) can be transformed as:
\[
\int_T f(x, y)\, dx\, dy = |J| \int_{\hat{T}} f(\mathbf{x}(\xi, \eta))\, d\xi\, d\eta,
\]
where \( |J| \) represents the absolute value of the determinant of the Jacobian matrix of the transformation:
\[
|J| = \left| (x_2 - x_1)(y_3 - y_1) - (x_3 - x_1)(y_2 - y_1) \right|.
\]

In benchmark tests, the relative error is as low as $2 \times 10^{-4}$ for logarithmic kernels 
\[
\ln(x^{2} + y^{2}) \quad \text{and} \quad \ln\big((x - 0.5)^{2} + (y - 0.5)^{2}\big)
\]
integrated over $[0,1] \times [0,1]$. To further validate the integration accuracy, we also performed tests with a variety of smooth, non-singular functions, confirming that the method consistently achieves high precision across both singular and regular integrands. Numerical results, compared against exact analytical integrals, confirm the robustness and reliability of the proposed scheme.

\subsubsection{Boundary Integration}

The boundary integrals are numerically evaluated via the trapezoidal rule, which provides a simple yet sufficiently accurate approximation for smooth integrands. The boundary is discretized into 400 points to facilitate this integration, with relative error measured using the $L^2$ norm.

\section{Numerical results}
\label{sec:numerical_results}
We conduct a systematic numerical investigation of three fundamental elliptic PDEs: the Laplace, Poisson, and Helmholtz equations, which are widely encountered in scientific and engineering applications. For Dirichlet boundary value problems, Eq. \ref{eq:laplace} simplifies to the case where $\partial\Omega_{D} = \partial \Omega$, meaning Dirichlet conditions are prescribed over the entire boundary. Our comprehensive evaluation examines MAD-BNO's performance across diverse scenarios, including both source-free and sourced equations with various boundary conditions and geometric configurations. \par

The test set is carefully designed to assess the generalization capability of the proposed method across the three PDE types. For the Laplace and Poisson boundary value problems, the test set comprises 50 analytical functions---including logarithmic-type functions derived from fundamental solutions, exponential-trigonometric product functions, non-linear polynomial functions, linear functions, and constant functions---of which 20 are sampled from within the training distribution and 30 are entirely out-of-distribution. \par

For Laplace boundary value problem, the test set consists of real-valued functions, with the following five functions used to evaluate the model's generalization performance:
\begin{equation}\label{eq:test-functions}
	\begin{aligned}
		u_1(x, y) &= \ln\big((x - m)^2 + (y - n)^2\big) &\quad& \text{(logarithmic kernel)} \\
		u_2(x, y) &= m(x^2 - y^2) + nxy                 && \text{(harmonic polynomial)} \\
		u_3(x, y) &= \sin(mx - t_1)\, e^{(mx - t_2)}    && \text{(composite function)} \\
		u_4(x, y) &= mx + ny                            && \text{(linear function)} \\
		u_5(x, y) &= x^3 - 3xy^2                        && \text{(harmonic polynomial)}
	\end{aligned}
\end{equation}
Here, the parameters \( m \) and \( n \) are uniformly sampled from \([-4, 4]\). For the logarithmic kernel \( u_1 \) (the only function included in the training set), \( m \) is specifically chosen such that the corresponding source point \((m, n)\) lies strictly outside the computational domain \(\Omega\) to avoid singularities. Among these functions, all others are entirely excluded from the training data, enabling rigorous evaluation of the model's extrapolation capability beyond the training distribution. \par

In the case of Helmholtz equations, the test set consists of real-valued functions outside the training distribution, specifically in the following forms:
\begin{equation}
	u(x, y) = \sin(ax - p_1)\sin(by - p_2)
\end{equation}
and
\begin{equation}
	u(x, y) = \sin(cx - q_1)\sinh(dy - q_2),
\end{equation}
where the wave number $k$ satisfies:
\begin{equation}
	k^2 = a^2 + b^2 = c^2 - d^2.
\end{equation}
Here, $p_1$, $p_2$, $q_1$, and $q_2$ are phase shift parameters that spatially translate the functions. Coefficients are constrained to ranges: $a, b \in (0, k]$, $c, d \in (0, 2k]$, with $c > d$ as an additional condition. This enables a broader range of wave patterns, and the diverse construction ensures a rigorous evaluation landscape, demonstrating MAD-BNO's robustness and adaptability in solving a broad class of boundary value problems.\par
To investigate the influence of network architectural choices on model accuracy, we evaluate the effect of different depth and width of the boundary encoder. Table \ref{tab:depth_width_loss} shows the impact of network architecture on MAD-BNO's performance for solving the Laplace boundary value problems. The results highlight that architectural simplicity aligns best with the problem's linear nature: a model with no hidden layers---directly mapping a 400-dimensional input to a 400-dimensional output---achieves comparable accuracy to deeper or wider networks, while significantly reducing computational overhead. This effectiveness stems from the fact that the Dirichlet-to-Neumann map for the Laplace equation is inherently linear, making additional layers or nonlinearities not only unnecessary but potentially detrimental. Therefore, throughout this work, all Dirichlet-Neumann mappings are implemented using the \([400,\,400]\) configuration, i.e., a single linear fully connected layer with 400 input and 400 output units, which provides a robust yet efficient architecture well aligned with the underlying mathematical structure.\par
Building on the identified architecture, we further compare MAD-BNO with MAD and PI-DeepONet for performance evaluation. The comparative results in Table \ref{result1} demonstrate that MAD-BNO achieves a favorable balance between efficiency and accuracy, outperforming the compared methods across critical metrics. In terms of computational efficiency, MAD-BNO reduces training time by 85\%–95\% compared to PI-DeepONet and 70\%–80\% compared to MAD across all tested equations. It is important to note that the reported training time refers exclusively to the training of normal derivative predictions and does not include the time for subsequent boundary integral evaluations. This drastic speedup stems from its boundary-centric design: by restricting training
to Dirichlet-Neumann boundary data pairs, it eliminates the need for computationally expensive full-domain sampling as in PI-DeepONet and MAD. Evaluating the solution over the entire domain requires boundary integral computations, which take approximately 20–30 ms, and volume integral computations for source terms, which take about 2 s.
In terms of accuracy, MAD-BNO consistently outperforms PI-DeepONet by 1–2 orders of magnitude in both training loss and test error, with its advantages becoming more pronounced in complex scenarios (e.g., high-wavenumber Helmholtz equations). While MAD achieves slightly lower test errors in simple cases (e.g., Laplace equation), MAD-BNO maintains robustness in challenging settings---such as $k=100$ Helmholtz problems---where PI-DeepONet and even MAD degrade significantly. This resilience arises from MAD-BNO's use of synthetic data generated from fundamental solutions, which exhibits high quality akin to error-free analytical solutions and inherently encodes the physical structure of the PDEs, ensuring more reliable capture of underlying dynamics than PI-DeepONet's physics-informed loss formulation.\par
Notably, MAD-BNO provides accurate predictions of normal derivatives ($\frac{\partial u}{\partial n}$) with relative errors ranging from $1.39\times 10^{-2}$ to $2.88\times 10^{-2}$, which in most cases surpass the accuracy of normal derivatives obtained via automatic differentiation in MAD and PI-DeepONet. This enables direct reconstruction of interior solutions via boundary integral formulations. While MAD occasionally achieves slightly lower interior $L^2$ errors on simpler problems, part of the MAD-BNO error originates from the  approximation in the boundary integral reconstruction. Nevertheless, these differences are generally minor, and MAD-BNO consistently demonstrates substantially higher efficiency and greater robustness, especially in high-frequency or geometrically complex regimes.

\begin{table}[H]
	\centering
	\small 
	\setlength{\tabcolsep}{7pt} 
	\renewcommand{\arraystretch}{1.1}  
	\begin{tabular}{|c|c|c|c|c|}
		\hline
		\diagbox{Hidden Layers}{Neurons} & 200 & 400 & 800 & Loss (0 hidden) \\
		\hline
		0 & - & - & - & \textbf{6.07E-3} \\
		\hline
		1 & 7.75E-3 & 8.25E-3 & 1.07E-2 & - \\
		\hline
		2 & 9.48E-3 & 1.21E-2 & 1.33E-2 & - \\
		\hline
		4 & 1.63E-2 & 1.25E-2 & 2.05E-2 & - \\
		\hline
	\end{tabular}
	\caption{Laplace equation: Relative $L^2$ error between MAD-BNO predictions and the exact solution $u(x, y)$, for different numbers of hidden layers and neurons per layer in the boundary encoder. The last column indicates the loss the network without hidden layer. As seen from the table, the model with 0 hidden layers achieves the smallest test relative $L^2$ error.}
	\label{tab:depth_width_loss}
\end{table}

\begin{flushleft}
\begin{table}[H]
\centering
\small
\setlength{\tabcolsep}{6pt} % 调整列间距
\renewcommand{\arraystretch}{0.9} % 增加行高
\begin{tabular}{c|c|c|c|c|c|c}
\toprule
 Equation& Method & Training Time  &Epoch& Training Loss&$\frac{\partial u}{\partial n}$ Error &Total Error \\
\midrule
\multirow{2}{*}{$\Delta u=0$} & MAD-BNO &\textbf{2.61h} &50000& \textbf{4.27e-7}&1.39e-2&6.07e-3  \\
&MAD&14.93h&50000&7.14e-7&\textbf{1.14e-2}&\textbf{3.21e-3}\\
&PI-DeepONet & 31.09h  &50000& 7.21e-5&1.13e-1& 8.52e-2 \\
\midrule
\multirow{2}{*}{$\Delta u=f$} & MAD-BNO & \textbf{2.61h} &50000& 4.27e-7&\textbf{1.39e-2}&9.32e-3  \\
&MAD&83.69h&50000&\textbf{1.34e-8}&2.67e-2&\textbf{4.47e-3}\\
& PI-DeepONet & 125.47h  &50000&8.83e-5&7.23e-2& 4.92e-2 \\
\midrule
\multirow{2}{*}{$\Delta u+u=0$} & MAD-BNO &\textbf{2.58h} &50000& 1.17e-6 &\textbf{2.07e-2}&\textbf{4.72e-3} \\
&MAD&14.81h&50000&\textbf{5.69e-7}&4.86e-2&5.34e-3\\
&PI-DeepONet & 34.55h &50000& 4.36e-5&7.67e-2& 2.57e-2 \\
\midrule                           
\multirow{2}{*}{$\Delta u+100u=0$} & MAD-BNO &\textbf{2.73h} &50000&\textbf{1.05e-6} &\textbf{2.49e-2}&\textbf{5.35e-3} \\
&MAD&15.32h&50000&3.27e-5&2.96e-2&5.87e-3\\
&PI-DeepONet & 34.50h &50000& 2.33e-2&1.00e+0& 1.00e+0 \\
\midrule                           
\multirow{2}{*}{$\Delta u+10000u=0$} & MAD-BNO &\textbf{4.12h} &50000&\textbf{ 1.26e-5} &\textbf{2.88e-2}&\textbf{2.73e-2} \\
  &     MAD   &   14.75h   &    50000   &    7.93e-2    &    1.26e+0     &     9.98e-1   \\
&     PI-DeepONet    &    34.29h   &    50000    &    2.31e-2     &    1.00e+0   &    1.00e+0     \\
\bottomrule
\end{tabular}
\caption{Performance comparison of MAD-BNO, MAD, and PI-DeepONet in operator learning for Laplace(Eq.~\ref{eq:laplace}), Poisson(Eq.~\ref{eq:poisson}), and Helmholtz(Eq.~\ref{eq:helmholtz}) Equations with Dirichlet boundary conditions.  \textbf{Total Error} refers to the average relative $L^2$ error between the interior solutions reconstructed via boundary integral and the corresponding exact solutions across multiple test cases.
}\label{result1}
\end{table}
\end{flushleft}

\subsection{Laplace equation}
\subsubsection{Dirichlet Boundary Value Problem}
\begin{figure}[H]
    \centering
    \includegraphics[width=0.8\linewidth]{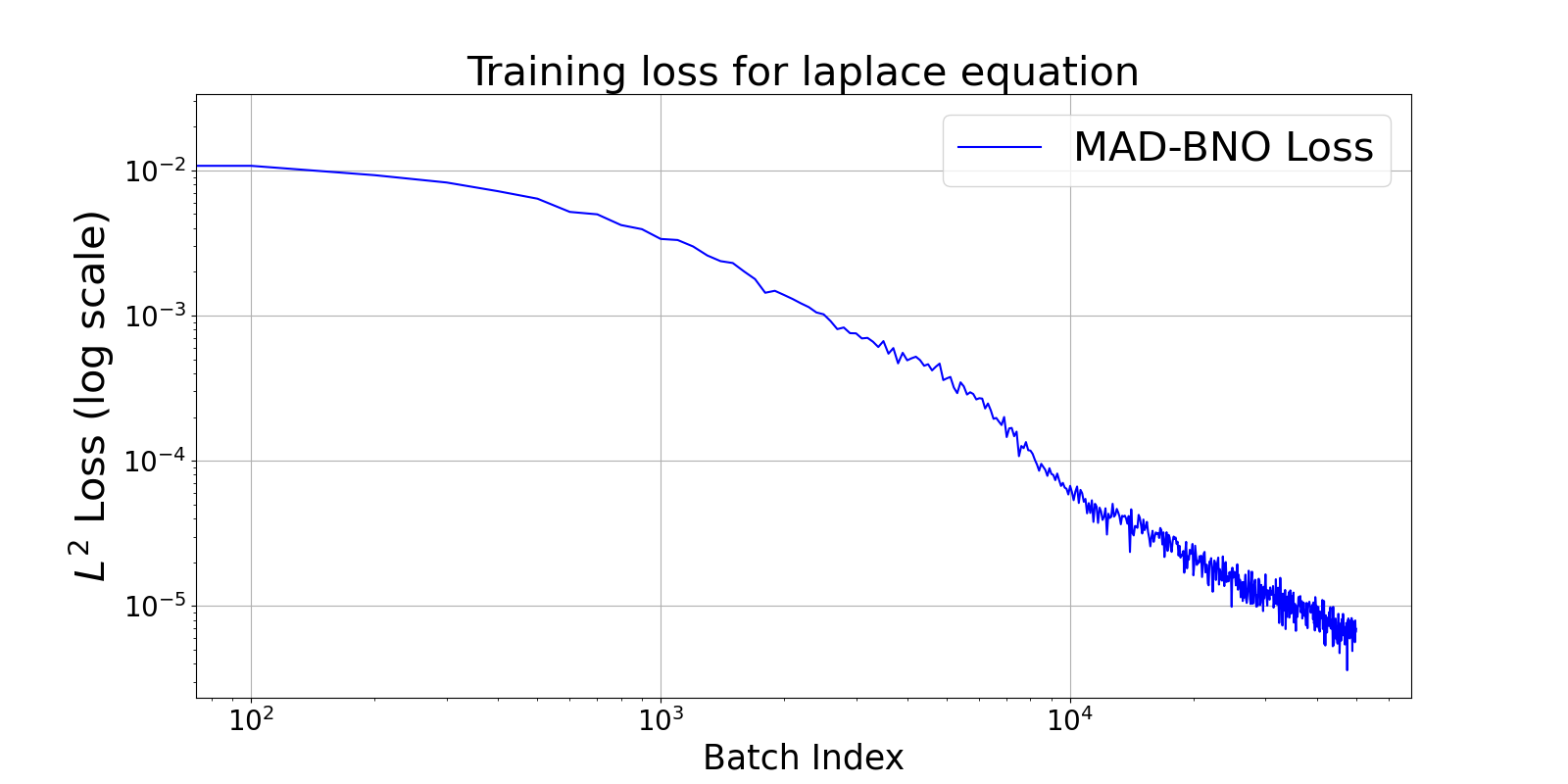}
    \caption{Training loss curve for the Laplace equation (Eq.~\ref{eq:laplace}) with Dirichlet boundary conditions. The vertical axis is shown on a logarithmic scale to emphasize the convergence rate of the model. The plotted loss values represent the average loss within every 100 epochs to better illustrate the training progress.
}\label{loss}
\end{figure}

\begin{figure}[H]
    \centering
    \begin{subfigure}[b]{0.3\textwidth}
        \centering
        \includegraphics[width=\textwidth]{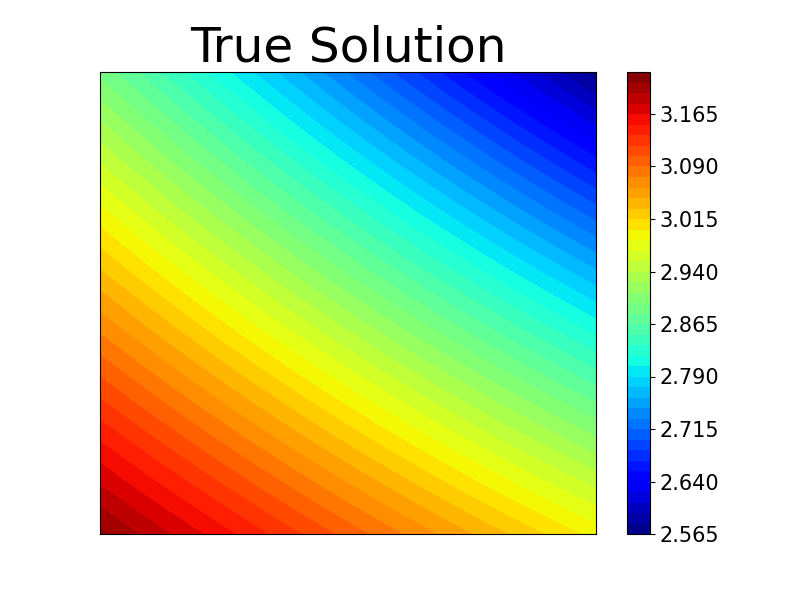}
        \caption{}
        \label{}
    \end{subfigure}
    \hfill
    \begin{subfigure}[b]{0.3\textwidth}
        \centering
        \includegraphics[width=\textwidth]{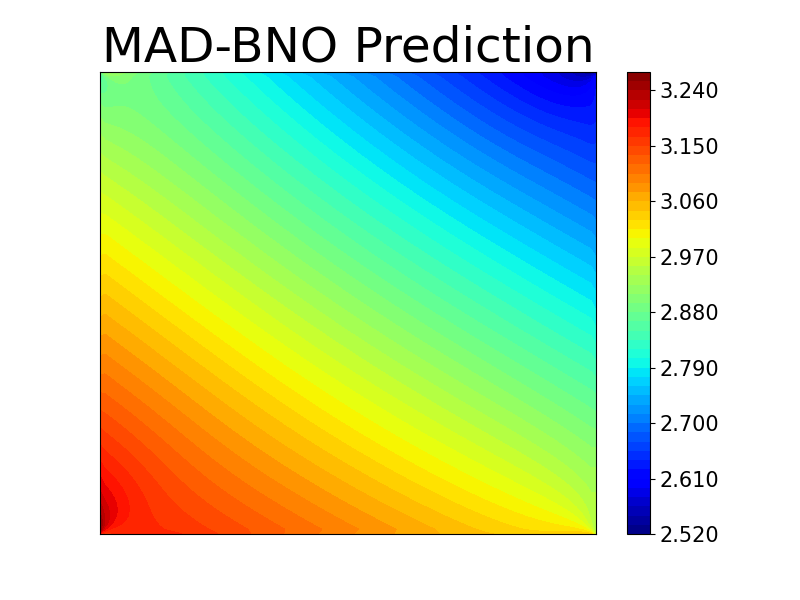}
        \caption{}
        \label{}
    \end{subfigure}
    \hfill
    \label{}
    \begin{subfigure}[b]{0.3\textwidth}
        \centering
        \includegraphics[width=\textwidth]{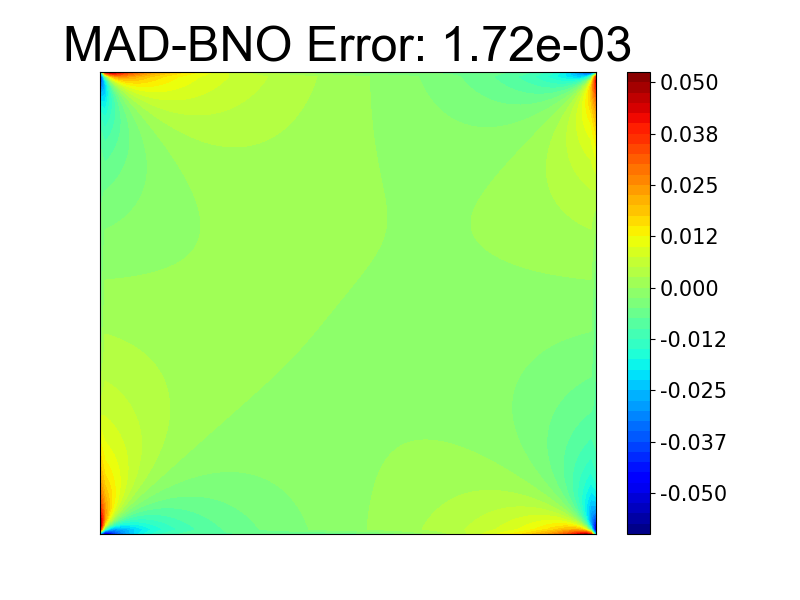}
        \caption{}
        \label{}
    \end{subfigure}
    \hfill
    \label{}
        \centering

    \begin{subfigure}[b]{0.3\textwidth}
        \centering
        \includegraphics[width=\textwidth]{real_sq.png}
        \caption{}
        \label{}
    \end{subfigure}
    \hfill
    \begin{subfigure}[b]{0.3\textwidth}
        \centering
        \includegraphics[width=\textwidth]{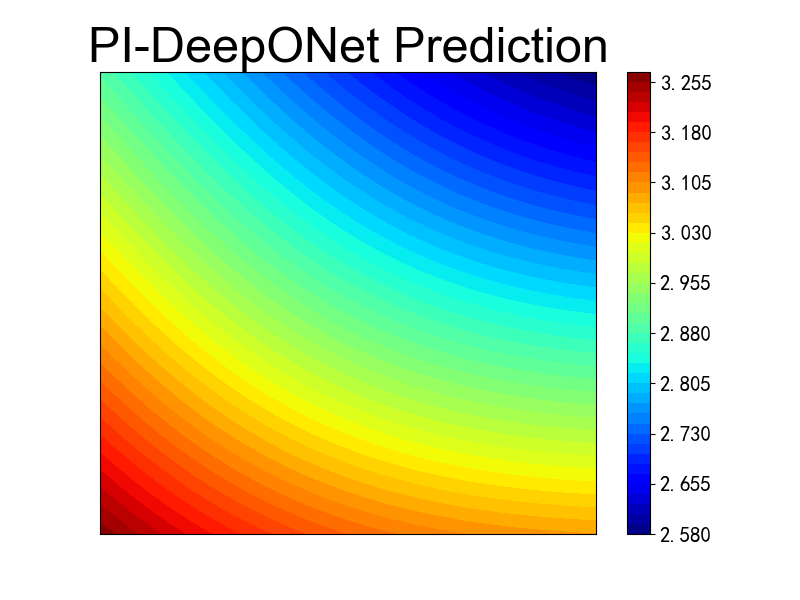}
        \caption{}
        \label{}
    \end{subfigure}
    \hfill
    \label{}
    \begin{subfigure}[b]{0.3\textwidth}
        \centering
        \includegraphics[width=\textwidth]{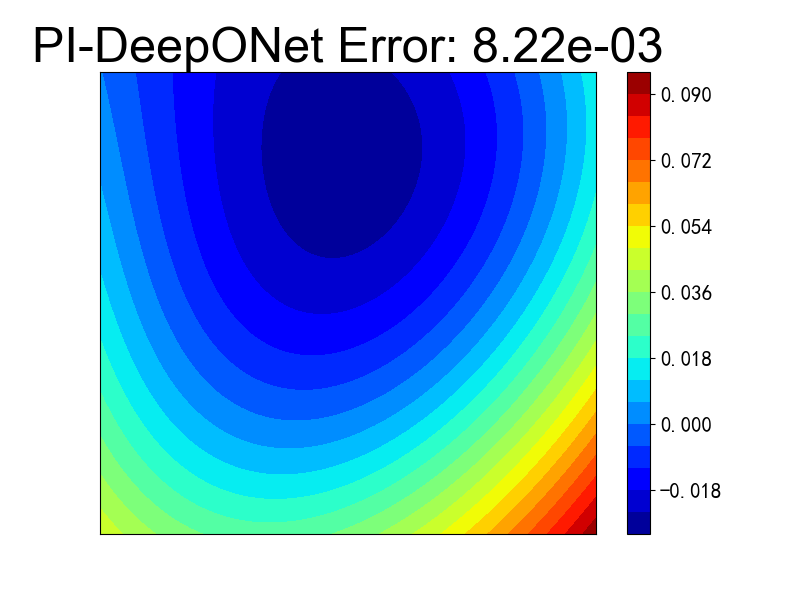}
        \caption{}
        \label{}
    \end{subfigure}
    \hfill
    \label{}
    		\begin{subfigure}[b]{1\textwidth}
    	\centering
    	\includegraphics[width=\textwidth]{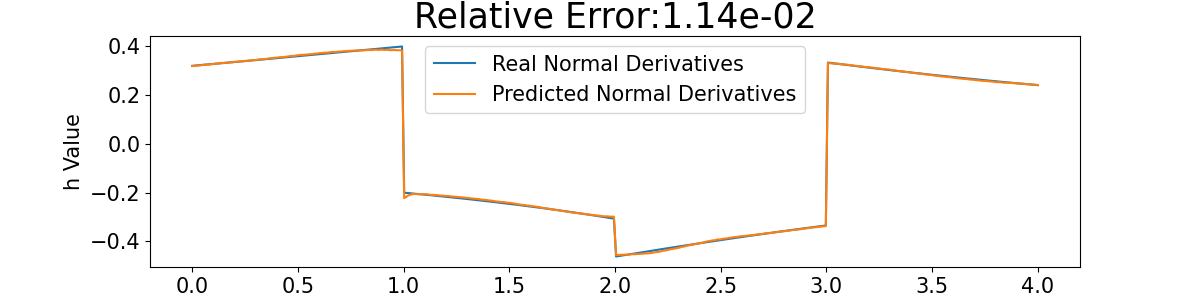}
    	\caption{}
    	\label{}
    \end{subfigure}
    \caption{Comparison of MAD-BNO and PI-DeepONet's predictions for the analytical solution $\bm{u(x, y) = \ln\left( (x - 3)^2 + (y - 4)^2 \right)}$ of the Laplace equation in the square domain $(0,1) \times (0,1)$. (\textbf{g}): Comparison of the Neumann boundary values predicted by MAD-BNO with the exact values, where the square boundary of the domain is parameterized into the interval $[0,4]$ by traversing counterclockwise from the point $(0,0)$.
    }\label{boundary 1 Laplace}
\end{figure}

\begin{figure}[H]
    \centering
    \begin{subfigure}[b]{0.3\textwidth}
        \centering
        \includegraphics[width=\textwidth]{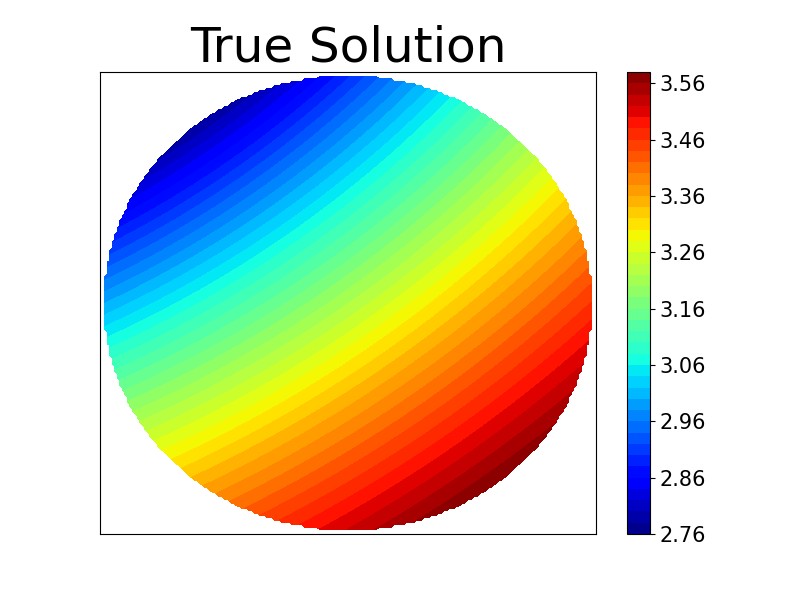}
        \caption{}
        \label{}
    \end{subfigure}
    \hfill
    \begin{subfigure}[b]{0.3\textwidth}
        \centering
        \includegraphics[width=\textwidth]{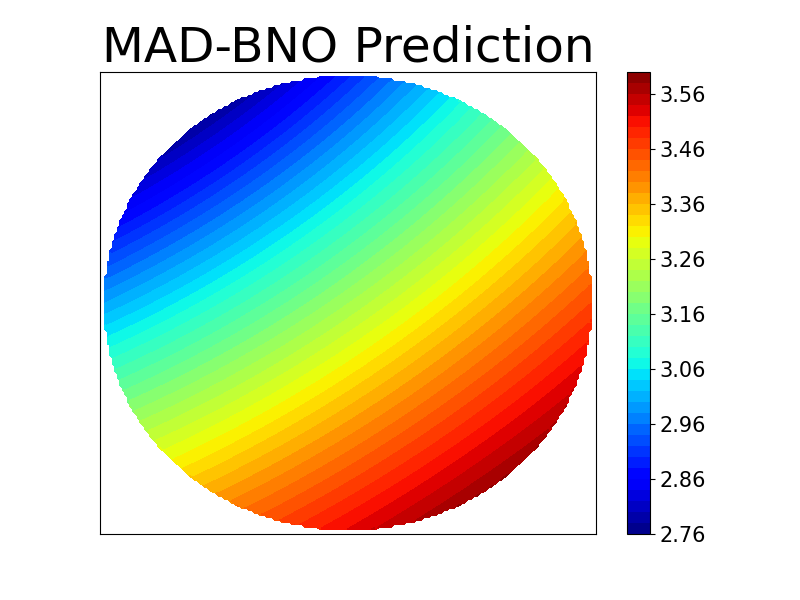}
        \caption{}
        \label{}
    \end{subfigure}
    \hfill
    \label{}
    \begin{subfigure}[b]{0.3\textwidth}
        \centering
        \includegraphics[width=\textwidth]{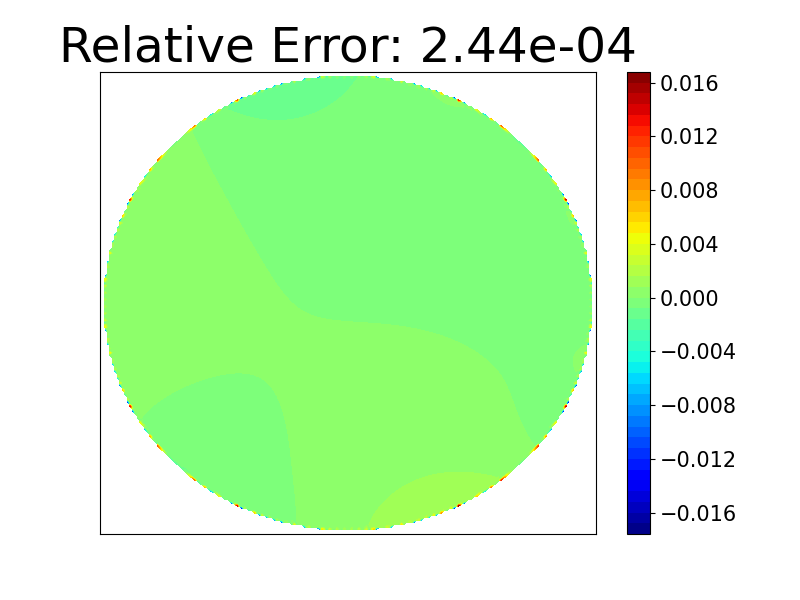}
        \caption{}
        \label{}
    \end{subfigure}
    \hfill
    \label{}
        \centering
    \begin{subfigure}[b]{0.3\textwidth}
        \centering
        \includegraphics[width=\textwidth]{real_1.png}
        \caption{}
        \label{}
    \end{subfigure}
    \hfill
    \begin{subfigure}[b]{0.3\textwidth}
        \centering
        \includegraphics[width=\textwidth]{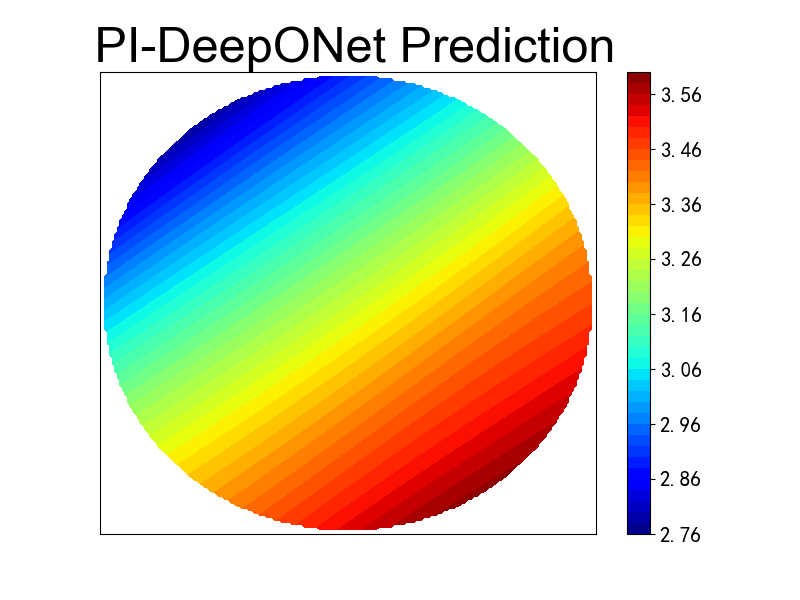}
        \caption{}
        \label{}
    \end{subfigure}
    \hfill
    \label{}
    \begin{subfigure}[b]{0.3\textwidth}
        \centering
        \includegraphics[width=\textwidth]{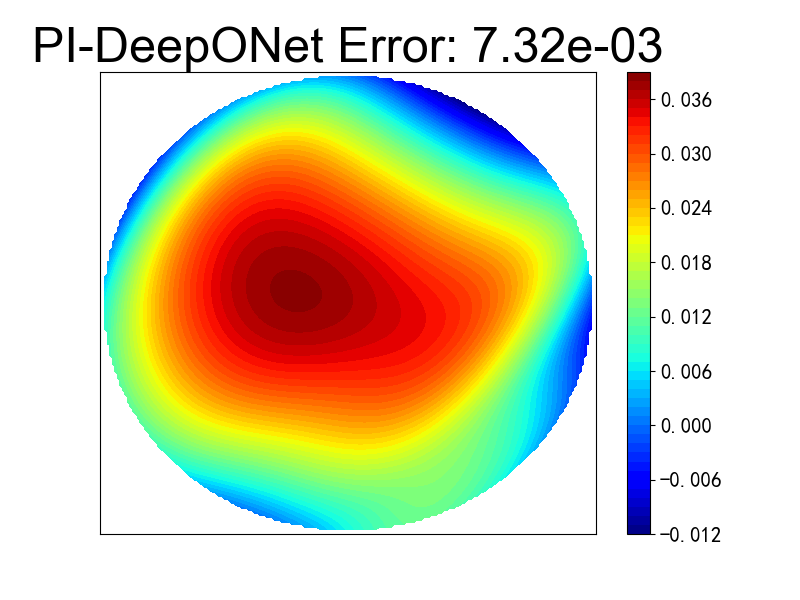}
        \caption{}
        \label{}
    \end{subfigure}
    \hfill
    \label{}
    \caption{Comparison of MAD-BNO and PI-DeepONet's predictions for the analytical solution $\bm {u(x, y) = \ln\left( (x + 3)^2 + (y - 4)^2 \right)} $ of the Laplace equation 
    	in a domain enclosed by the polar curve \( r(\theta) =1 \), 
    	with \( \theta \in [0, 2\pi) \).}\label{boundary 2 Laplace1}
\end{figure}

\begin{figure}[H]
    \centering
    \begin{subfigure}[b]{0.3\textwidth}
        \centering
        \includegraphics[width=\textwidth]{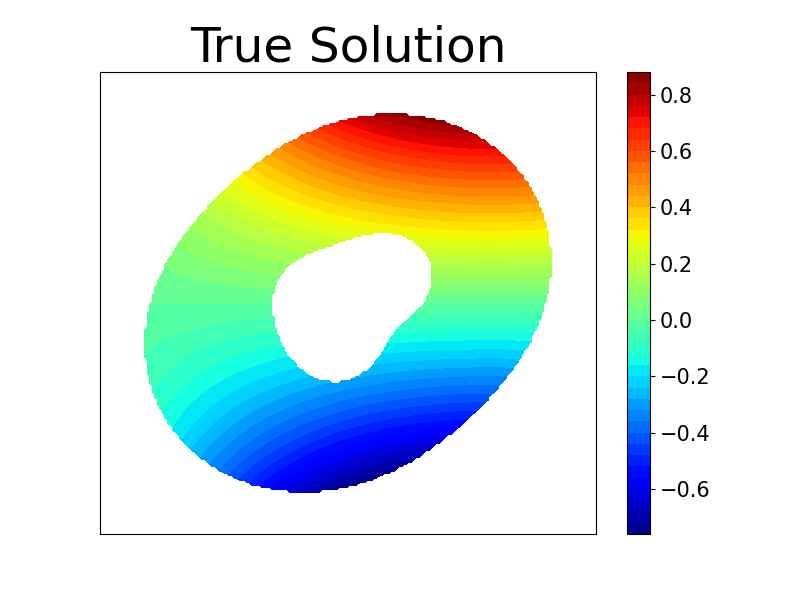}
        \caption{}
        \label{}
    \end{subfigure}
    \hfill
    \begin{subfigure}[b]{0.3\textwidth}
        \centering
        \includegraphics[width=\textwidth]{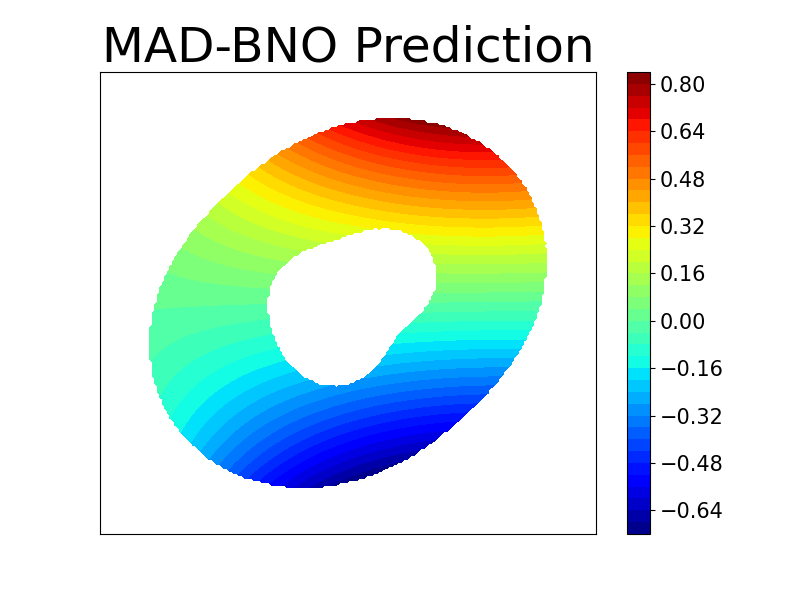}
        \caption{}
        \label{}
    \end{subfigure}
    \hfill
    \label{}
    \begin{subfigure}[b]{0.3\textwidth}
        \centering
        \includegraphics[width=\textwidth]{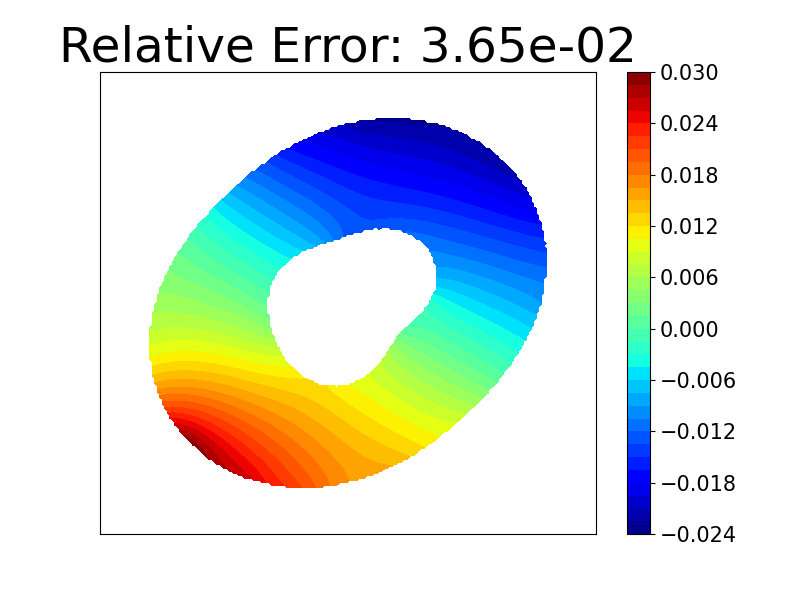}
        \caption{}
        \label{}
    \end{subfigure}
    \hfill
    \label{}
        \centering
    \begin{subfigure}[b]{0.3\textwidth}
        \centering
        \includegraphics[width=\textwidth]{sinsinh1_real.png}
        \caption{}
        \label{}
    \end{subfigure}
    \hfill
    \begin{subfigure}[b]{0.3\textwidth}
        \centering
        \includegraphics[width=\textwidth]{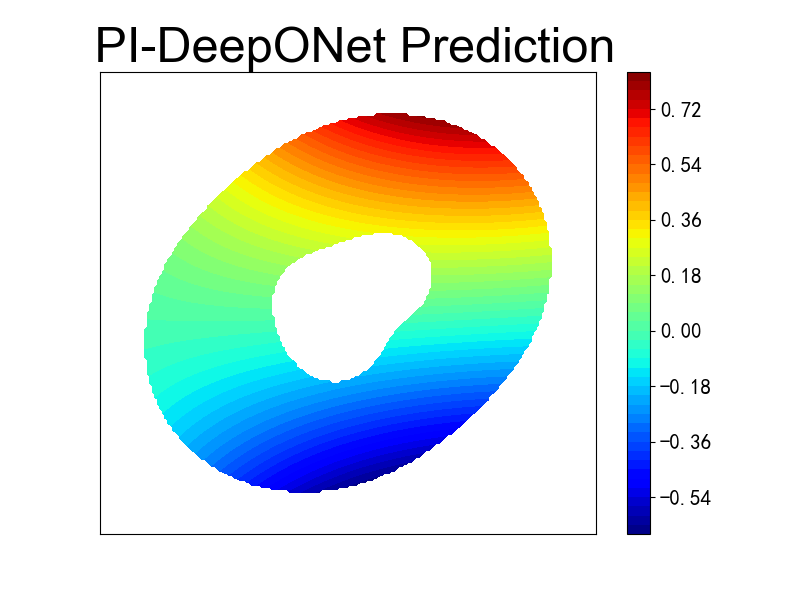}
        \caption{}
        \label{}
    \end{subfigure}
    \hfill
    \label{}
    \begin{subfigure}[b]{0.3\textwidth}
        \centering
        \includegraphics[width=\textwidth]{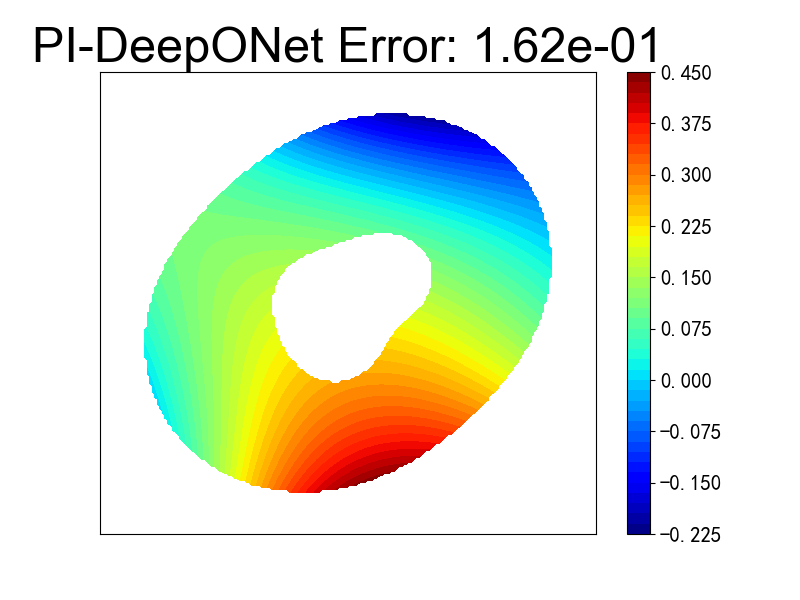}
        \caption{}
        \label{}
    \end{subfigure}
    \hfill
    \label{}
    \caption{Comparison of MAD-BNO and PI-DeepONet's predictions for the analytical solution  $\bm{ u(x, y) =sin(x+1)sinh(y) }$ of the Laplace equation 
    	in a domain enclosed by the polar curve \( r_{1}(\theta) = 0.8+0.1sin(2\theta) \) and \( r_{2}(\theta) = 0.3+0.05sin(2\theta)+0.03sin(3\theta) \), 
    	with \( \theta \in [0, 2\pi) \).
 }\label{boundary 2 Laplace2}
\end{figure}

\begin{figure}[H]
    \centering
    \begin{subfigure}[b]{0.3\textwidth}
        \centering
        \includegraphics[width=\textwidth]{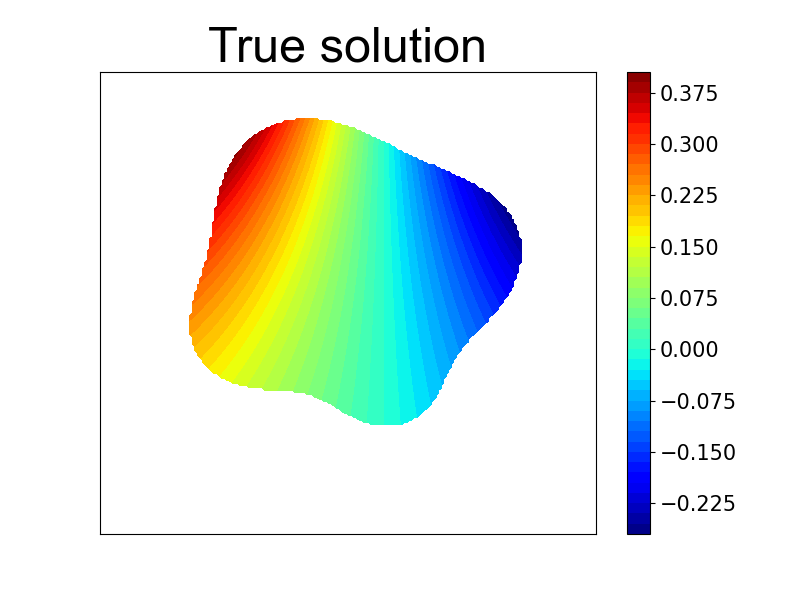}
        \caption{}
        \label{=}
    \end{subfigure}
    \hfill
    \begin{subfigure}[b]{0.3\textwidth}
        \centering
        \includegraphics[width=\textwidth]{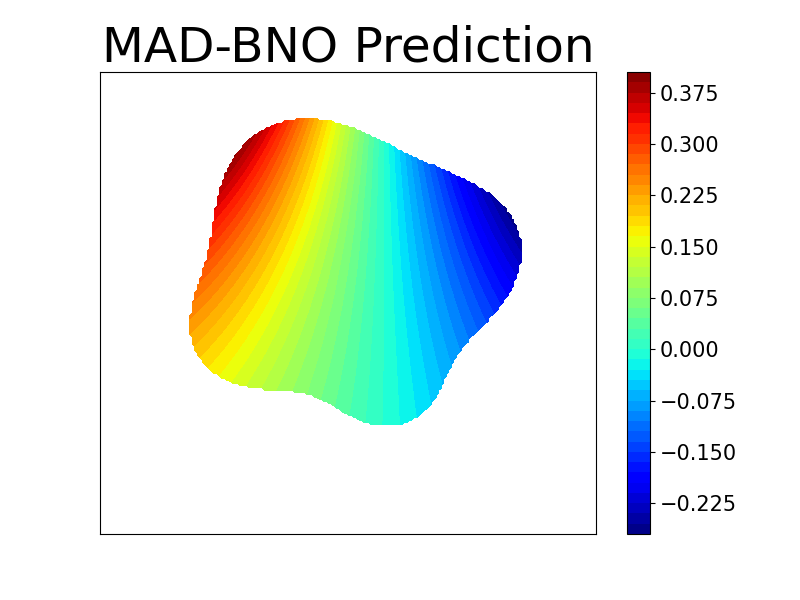}
        \caption{}
        \label{=}
    \end{subfigure}
    \hfill
    \label{}
    \begin{subfigure}[b]{0.3\textwidth}
        \centering
        \includegraphics[width=\textwidth]{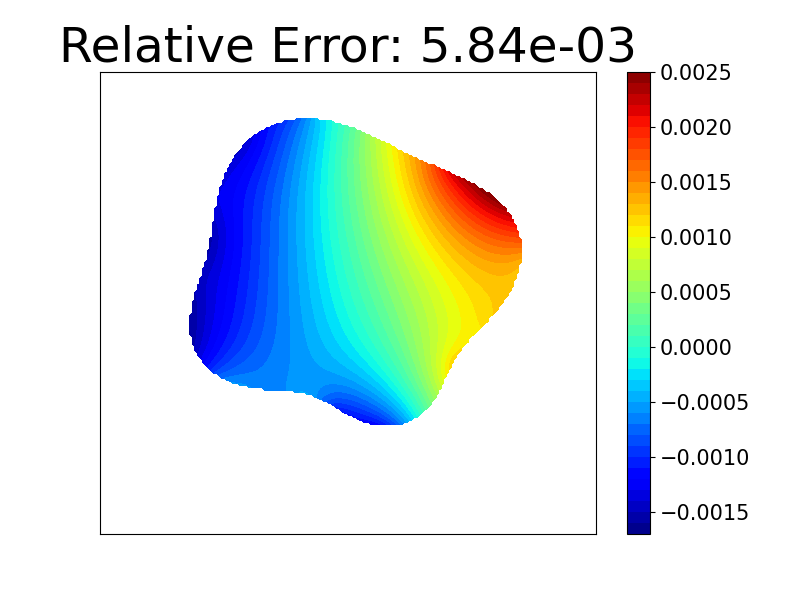}
        \caption{}
        \label{}
    \end{subfigure}
    \hfill
    \label{}
        \centering
    \begin{subfigure}[b]{0.3\textwidth}
        \centering
        \includegraphics[width=\textwidth]{real_2.png}
        \caption{}
        \label{}
    \end{subfigure}
    \hfill
    \begin{subfigure}[b]{0.3\textwidth}
        \centering
        \includegraphics[width=\textwidth]{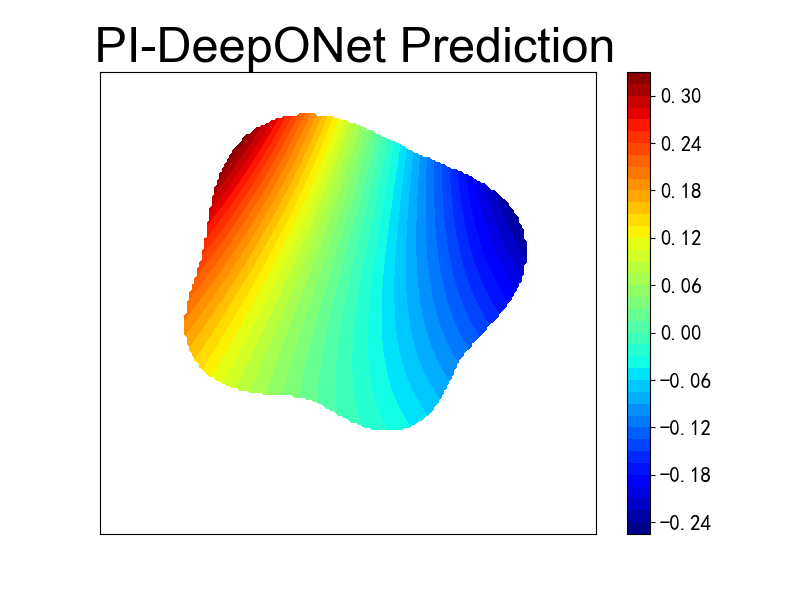}
        \caption{}
        \label{}
    \end{subfigure}
    \hfill
    \label{}
    \begin{subfigure}[b]{0.3\textwidth}
        \centering
        \includegraphics[width=\textwidth]{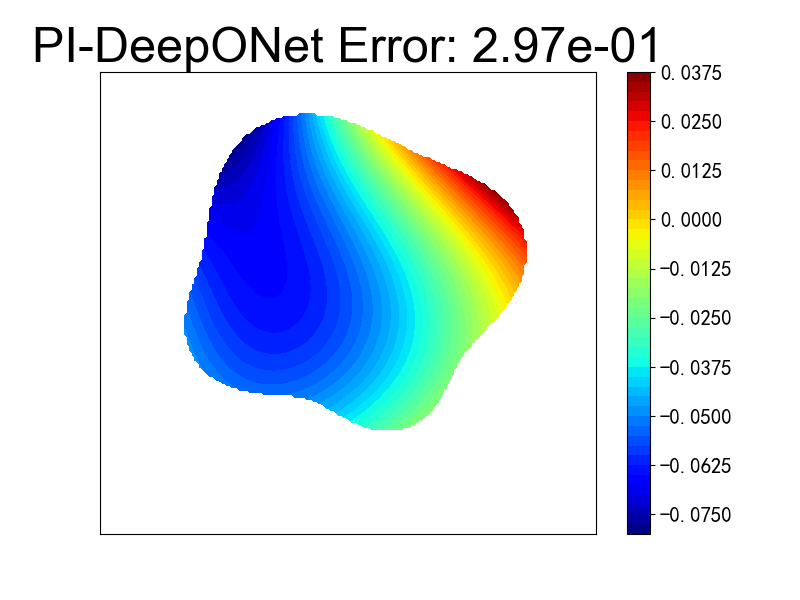}
        \caption{}
        \label{}
    \end{subfigure}
    \hfill 
    \label{}
    \caption{Comparison of MAD-BNO and PI-DeepONet's predictions for the analytical solution  $ \bm{u(x, y) = sin(x+3)e^{y-1}} $ of the Laplace equation 
    	in a domain enclosed by the polar curve \( r(\theta) = 0.65 + 0.14\sin\theta + 0.07\sin(4\theta) \), 
    	with \( \theta \in [0, 2\pi) \).}.
\label{boundary 3 Laplace1}
\end{figure}

\begin{figure}[H]
    \centering
    \begin{subfigure}[b]{0.3\textwidth}
        \centering
        \includegraphics[width=\textwidth]{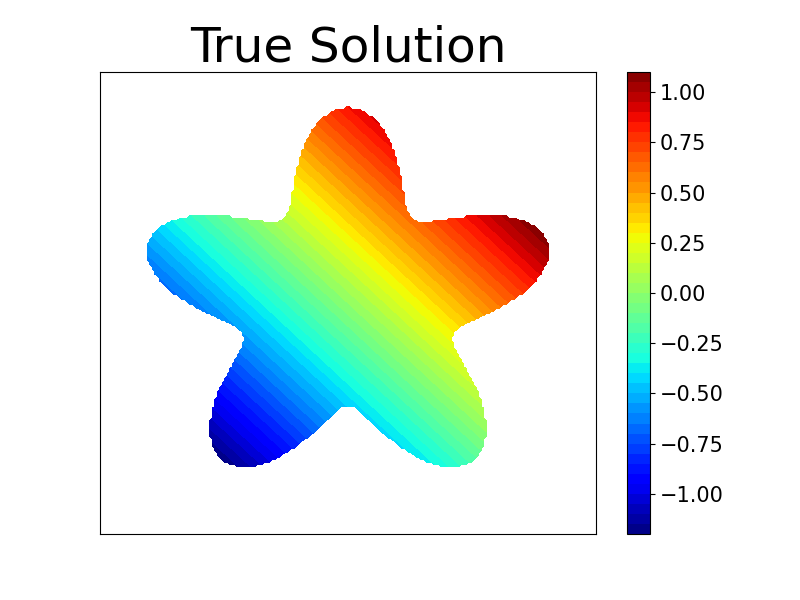}
        \caption{}
        \label{}
    \end{subfigure}
    \hfill
    \begin{subfigure}[b]{0.3\textwidth}
        \centering
        \includegraphics[width=\textwidth]{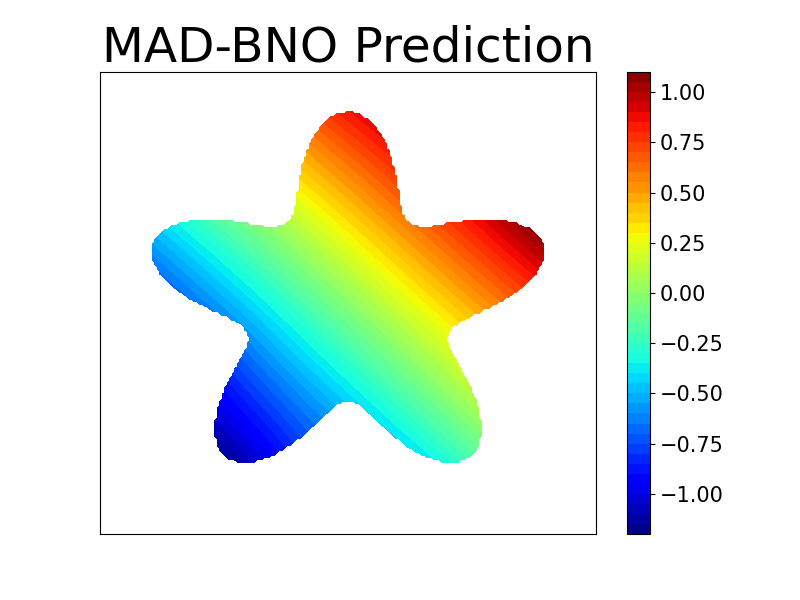}
        \caption{}
        \label{}
    \end{subfigure}
    \hfill
    \label{}
    \begin{subfigure}[b]{0.3\textwidth}
        \centering
        \includegraphics[width=\textwidth]{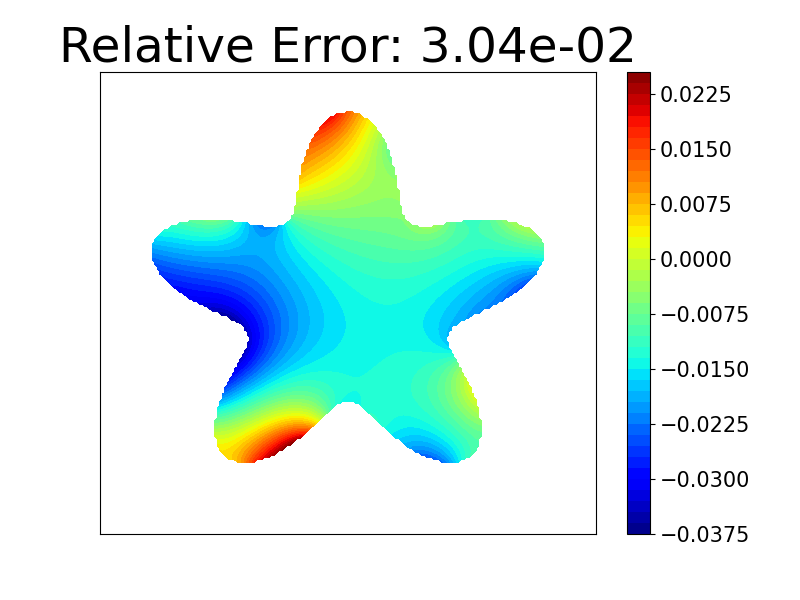}
        \caption{}
        \label{}
    \end{subfigure}
    \hfill
    \label{}
        \centering
    \begin{subfigure}[b]{0.3\textwidth}
        \centering
        \includegraphics[width=\textwidth]{linear_real.png}
        \caption{}
        \label{}
    \end{subfigure}
    \hfill
    \begin{subfigure}[b]{0.3\textwidth}
        \centering
        \includegraphics[width=\textwidth]{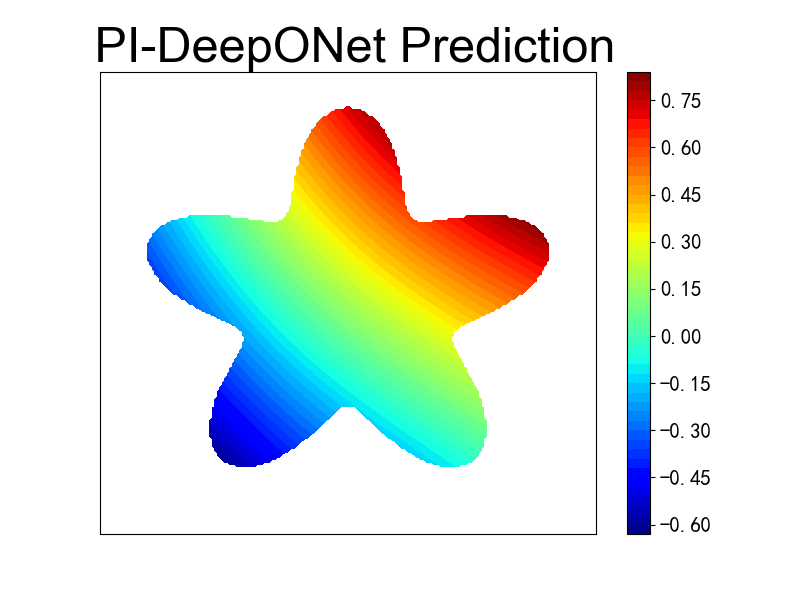}
        \caption{}
        \label{}
    \end{subfigure}
    \hfill
    \label{}
    \begin{subfigure}[b]{0.3\textwidth}
        \centering
        \includegraphics[width=\textwidth]{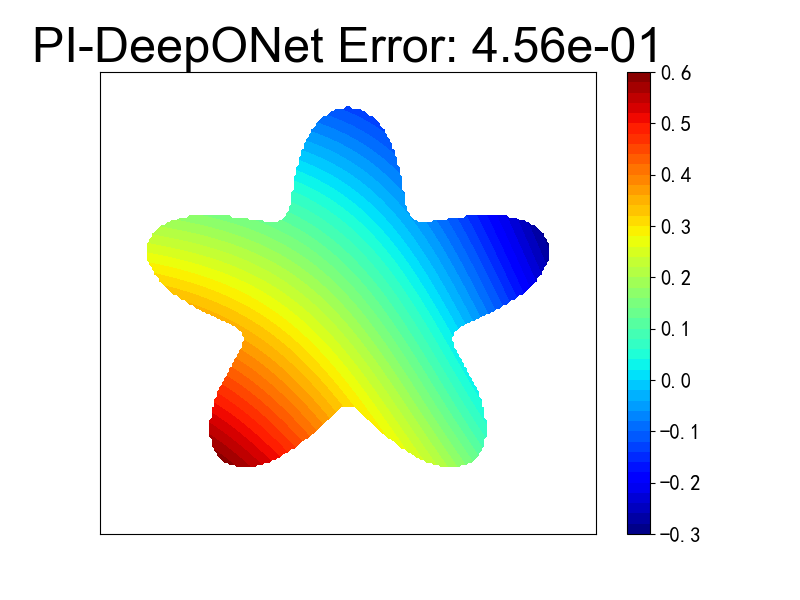}
        \caption{}
        \label{}
    \end{subfigure}
    \hfill
    \label{}
    \caption{Comparison of MAD-BNO and PI-DeepONet's predictions for the analytical solution   $\bm{ u(x, y) = x+y }$ of the Laplace equation 
    	in a domain enclosed by the polar curve \( r(\theta) = 0.65 +  0.2\sin(5\theta) \), 
    	with \( \theta \in [0, 2\pi) \).}.
\label{boundary 3 Laplace2 }
\end{figure}

\begin{figure}[H]
    \centering
    \begin{subfigure}[b]{0.3\textwidth}
        \centering
        \includegraphics[width=\textwidth]{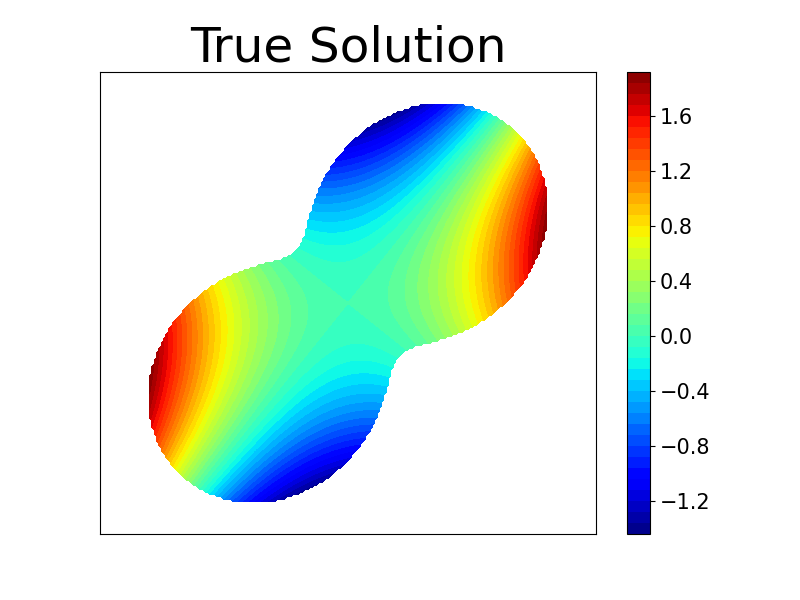}
        \caption{}
        \label{}
    \end{subfigure}
    \hfill
    \begin{subfigure}[b]{0.3\textwidth}
        \centering
        \includegraphics[width=\textwidth]{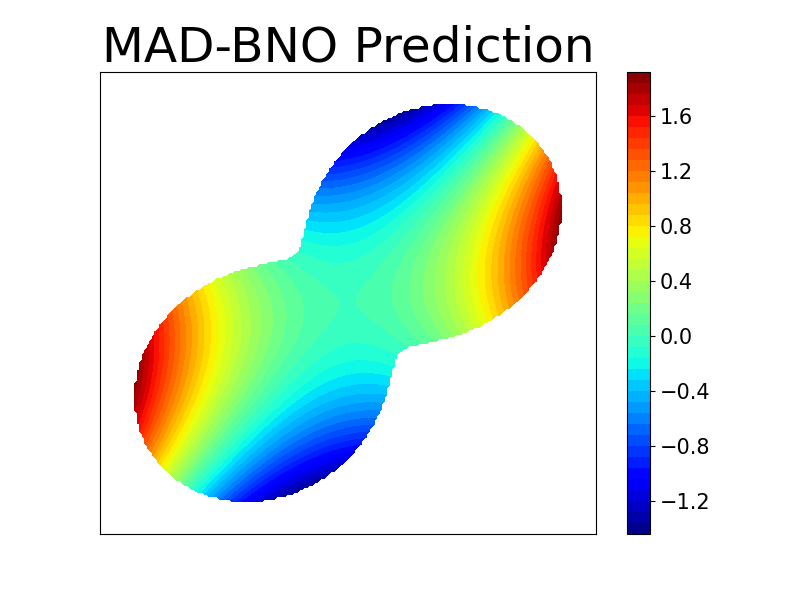}
        \caption{}
        \label{}
    \end{subfigure}
    \hfill
    \label{}
    \begin{subfigure}[b]{0.3\textwidth}
        \centering
        \includegraphics[width=\textwidth]{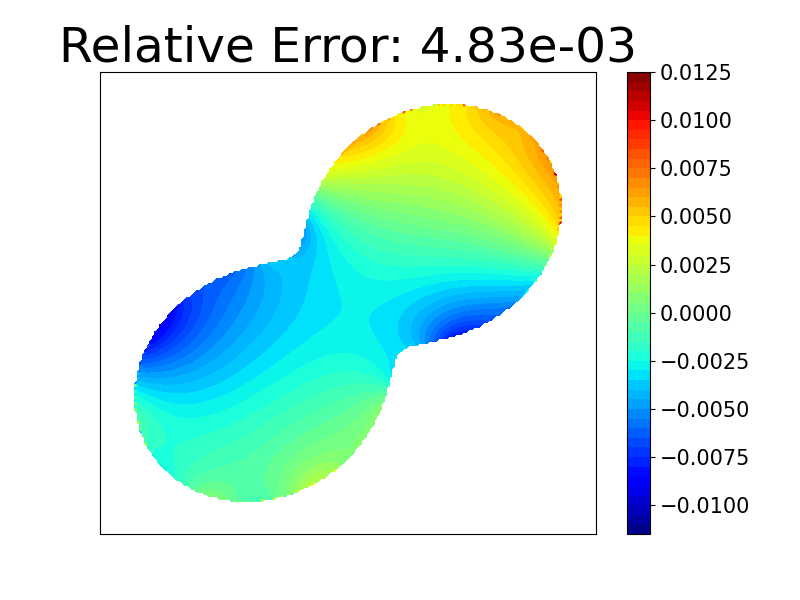}
        \caption{}
        \label{}
    \end{subfigure}
    \hfill
    \label{}
        \centering
    \begin{subfigure}[b]{0.3\textwidth}
        \centering
        \includegraphics[width=\textwidth]{real_3.png}
        \caption{}
        \label{}
    \end{subfigure}
    \hfill
    \begin{subfigure}[b]{0.3\textwidth}
        \centering
        \includegraphics[width=\textwidth]{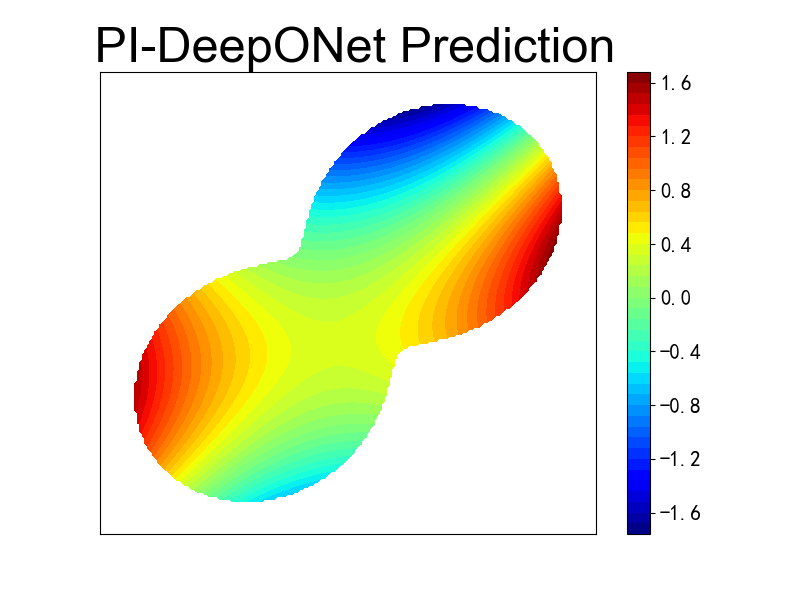}
        \caption{}
        \label{}
    \end{subfigure}
    \hfill
    \label{}
    \begin{subfigure}[b]{0.3\textwidth}
        \centering
        \includegraphics[width=\textwidth]{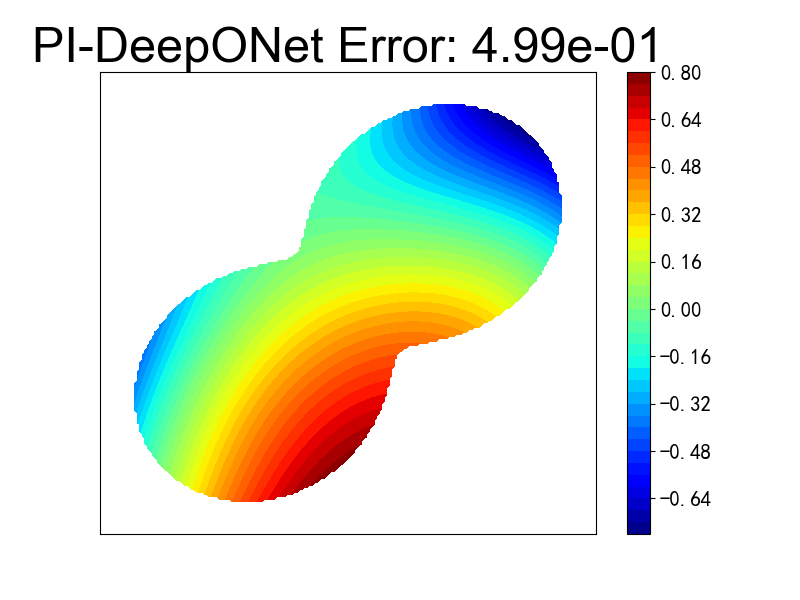}
        \caption{}
        \label{}
    \end{subfigure}
    \hfill
    \label{}
    \caption{Comparison of MAD-BNO and PI-DeepONet's predictions for the analytical solution  $ \bm{u(x, y) = 2.5(x^{2}-y^{2})+1.125xy} $ of the Laplace equation 
    	in a domain enclosed by the polar curve \( r(\theta) = 0.7 + 0.35\sin(2\theta) \), 
    	with \( \theta \in [0, 2\pi) \).}.
\label{boundary 4 Laplace1}
\end{figure}

To further evaluate the training behavior and predictive performance of MAD-BNO, we examine both its training behavior and numerical solution quality. Fig.~\ref{loss} shows the training loss of MAD-BNO over epochs, demonstrating stable convergence. Figs. \ref{boundary 1 Laplace}-\ref{boundary 4 Laplace1} present numerical prediction results for four representative analytical solution forms, including logarithmic, exponential, linear, and polynomial types. Predictions are presented on a 100 $\times$ 100 grid of points. MAD-BNO consistently achieves higher prediction accuracy across all test scenarios than PI-DeepONet. These results highlight the robustness and efficiency of MAD-BNO in solving the Laplace equation. All errors in this paper are measured using the relative $L^{2}$ norm.




\subsubsection{Mixed Boundary Value Problem}
To further evaluate the versatility of MAD-BNO, we apply it to mixed boundary value problems, where segments of the boundary are subject to Dirichlet conditions and the others to Neumann conditions. This setup introduces additional complexity due to the coexistence of distinct boundary behaviors.\par
\begin{figure}[H]
	\centering
	\includegraphics[width=0.5\linewidth]{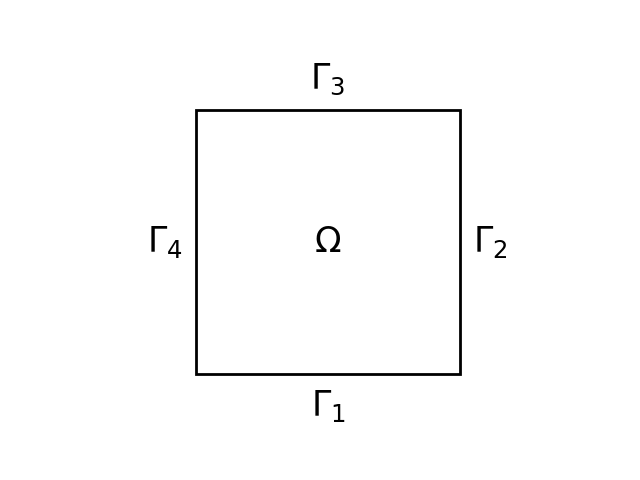}
	\caption{\textbf{Boundary Illustration}:
		The boundary of the square domain is divided into four segments. Each segment is assigned either exact Dirichlet or Neumann boundary conditions, allowing flexible combinations of boundary types along the boundary.}
	\label{square4}
\end{figure}

  
﻿
For the mixed boundary value problem experiment, we partitioned the boundary of the square domain \((0,1) \times (0,1)\) into four segments. Each segment is assigned either Dirichlet or Neumann conditions. The partitioning scheme is illustrated in Fig.~\ref{square4}. The model is trained to predict complementary boundary data for each segment given partial boundary information---this setup aims to assess its ability to recover missing boundary conditions and generalize across diverse mixed boundary scenarios.\par
\begin{figure}[H]\label{DN figure1}
	\centering
	\begin{subfigure}[b]{0.3\textwidth}
		\centering
		\includegraphics[width=\textwidth]{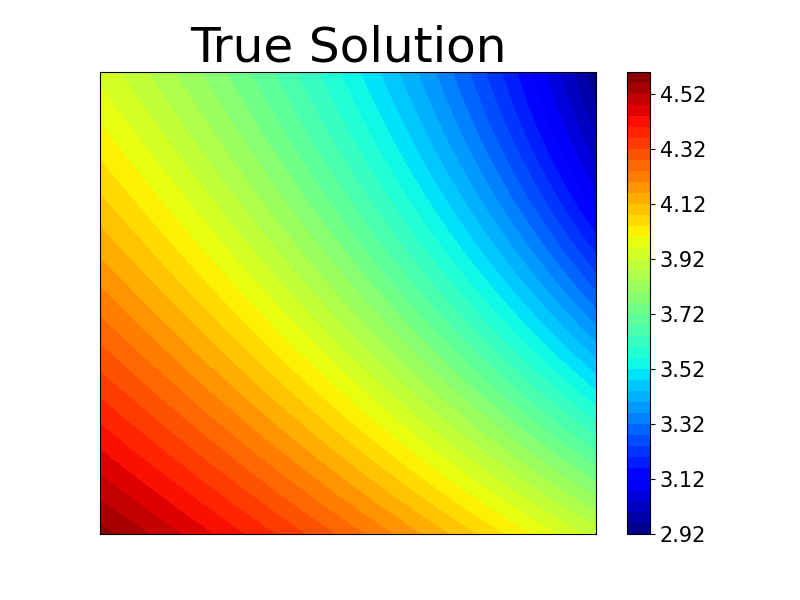}
		\caption{}
		\label{}
	\end{subfigure}
	\hfill
	\begin{subfigure}[b]{0.3\textwidth}
		\centering
		\includegraphics[width=\textwidth]{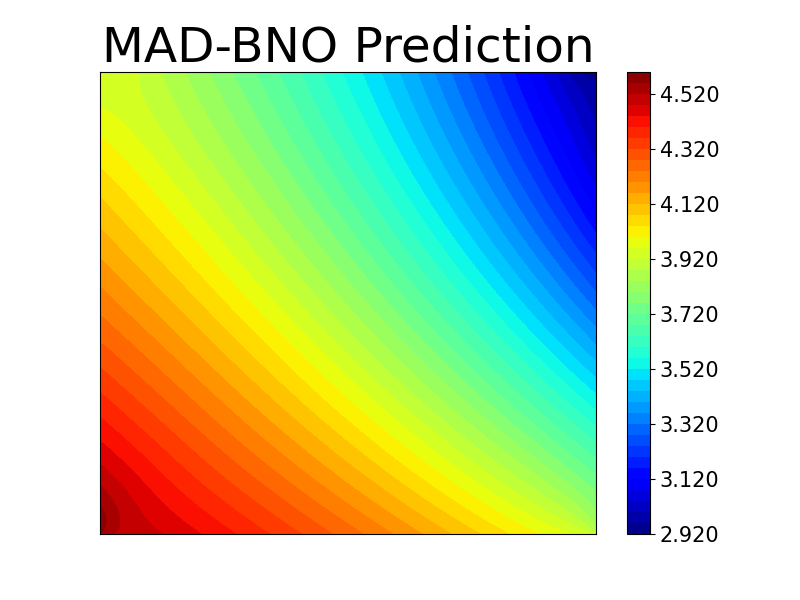}
		\caption{}
		\label{}
	\end{subfigure}
	\hfill
	\label{}
	\begin{subfigure}[b]{0.3\textwidth}
		\centering
		\includegraphics[width=\textwidth]{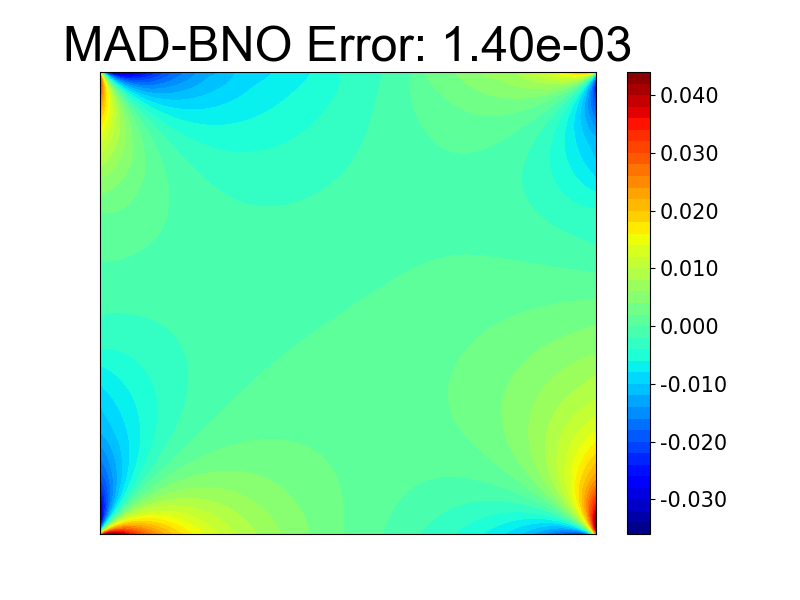}
		\caption{}
		\label{}
	\end{subfigure}
	\hfill
	
	\label{}
	\begin{subfigure}[b]{1\textwidth}
		\centering
		\includegraphics[width=\textwidth]{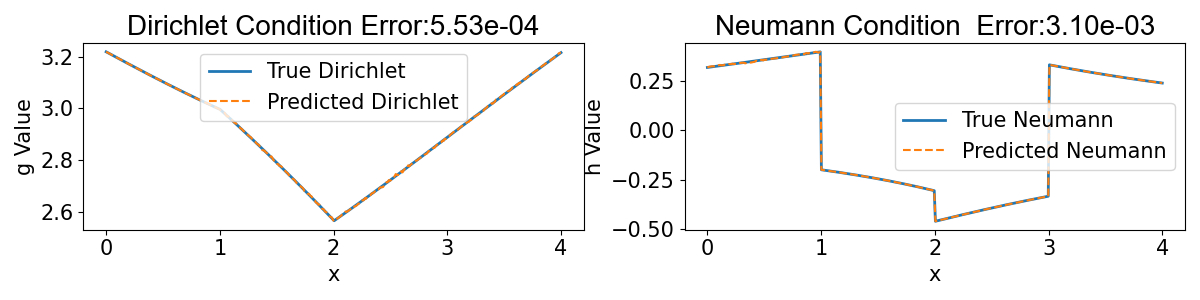}
		\caption{}
		\label{}
	\end{subfigure}
	\label{}
	\caption{Comparison of MAD-BNO and PI-DeepONet's predictions for the analytical solution $\bm{ u(x, y) = \ln\left( (x - 3)^2 + (y - 4)^2 \right) }$ of the Laplace equation in the square domain $(0,1) \times (0,1)$ with $\partial \Omega_{D}=\Gamma_{1}$ in (Fig.~\ref{square4}). (\textbf{d}):  Comparison of the Dirichlet and Neumann boundary values predicted by MAD-BNO with the exact values, where the square boundary of the domain is parameterized into the interval $[0,4]$ by traversing counterclockwise from the point $(0,0)$.\\
}\label{DN1}
\end{figure}

\begin{figure}[H]
	\centering
	\begin{subfigure}[b]{0.3\textwidth}
		\centering
		\includegraphics[width=\textwidth]{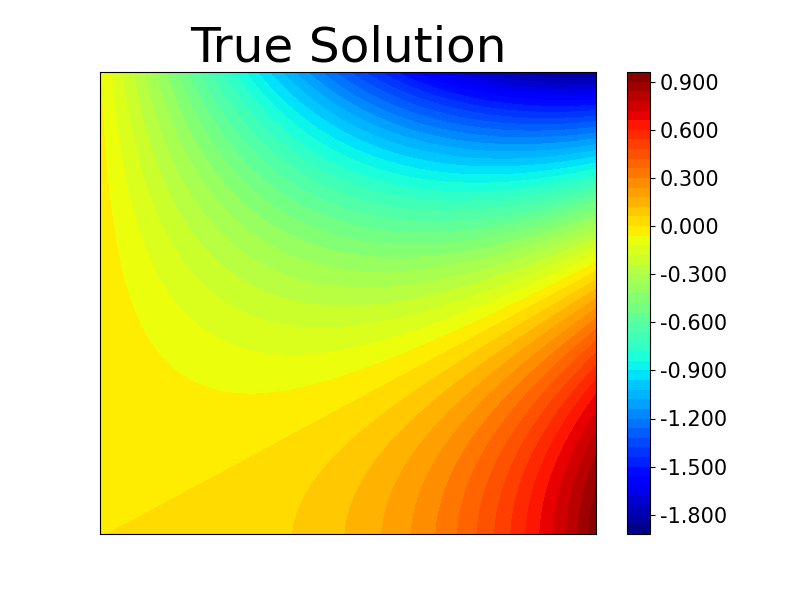}
		\caption{}
		\label{}
	\end{subfigure}
	\hfill
	\begin{subfigure}[b]{0.3\textwidth}
		\centering
		\includegraphics[width=\textwidth]{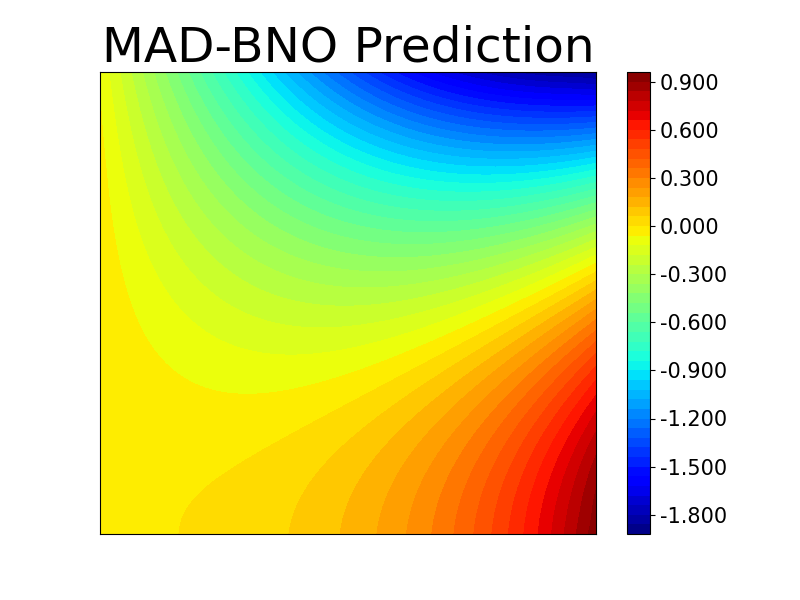}
		\caption{}
		\label{}
	\end{subfigure}
	\hfill
	\label{}
	\begin{subfigure}[b]{0.3\textwidth}
		\centering
		\includegraphics[width=\textwidth]{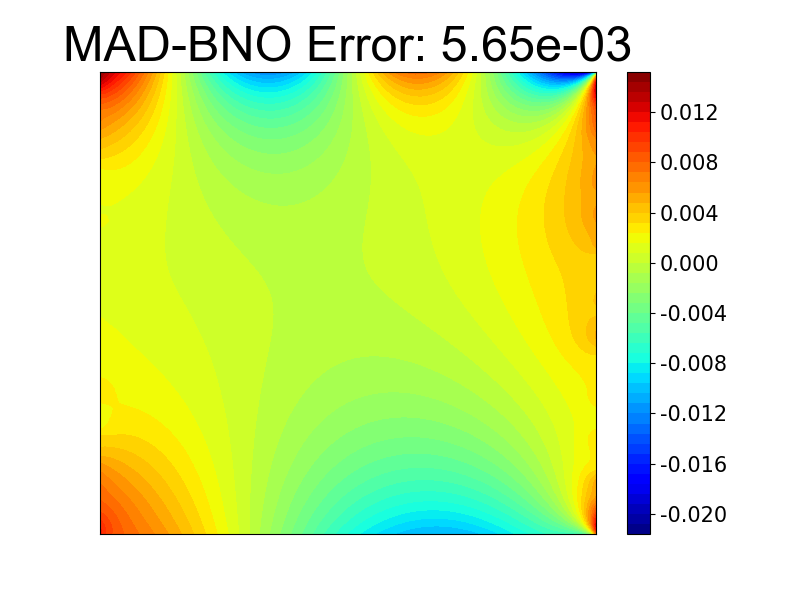}
		\caption{}
		\label{}
	\end{subfigure}
	\hfill
	\begin{subfigure}[b]{1\textwidth}
		\centering
		\includegraphics[width=\textwidth]{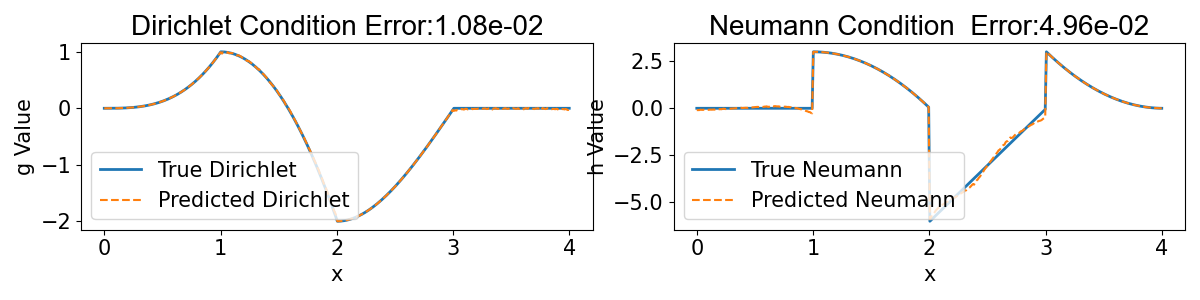}
		\caption{}
		\label{}
	\end{subfigure}
	\label{}
	\caption{Comparison of MAD-BNO and PI-DeepONet's predictions for the analytical solution  $\bm{u(x, y) = x^{3}-3xy^{2}} $ of the Laplace equation with $\partial \Omega_{D}=\Gamma_{1}\cup\Gamma_{3}$ in (Fig.~\ref{square4}). 
}\label{DN2}
\end{figure}

\begin{figure}[H]
	\centering
	\begin{subfigure}[b]{0.3\textwidth}
		\centering
		\includegraphics[width=\textwidth]{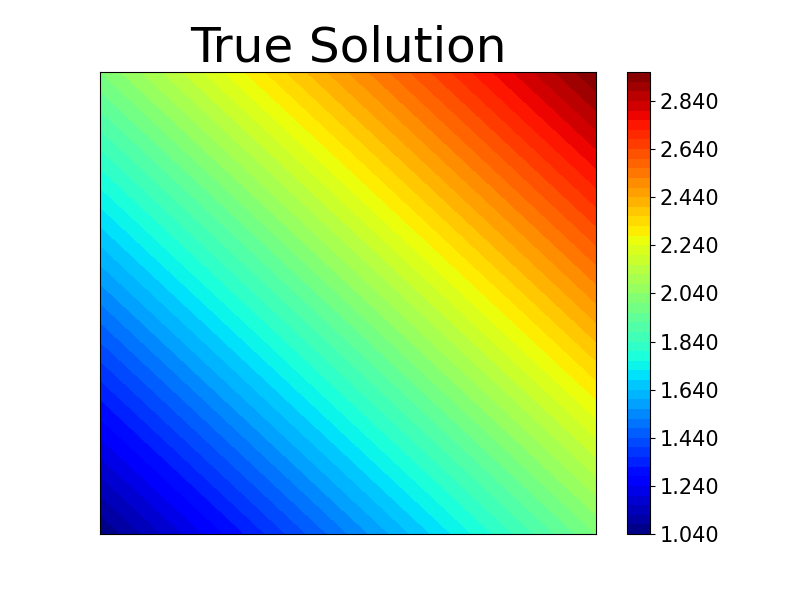}
		\caption{}
		\label{}
	\end{subfigure}
	\hfill
	\begin{subfigure}[b]{0.3\textwidth}
		\centering
		\includegraphics[width=\textwidth]{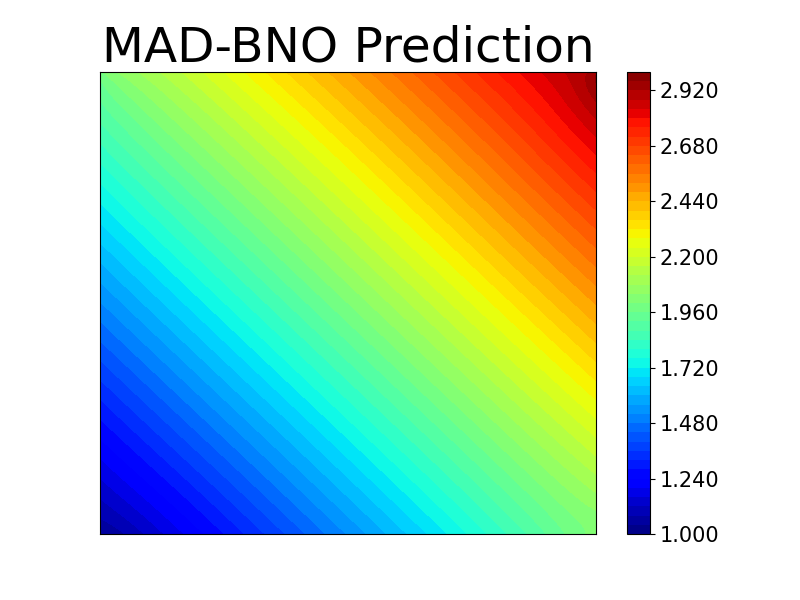}
		\caption{}
		\label{}
	\end{subfigure}
	\hfill
	\label{}
	\begin{subfigure}[b]{0.3\textwidth}
		\centering
		\includegraphics[width=\textwidth]{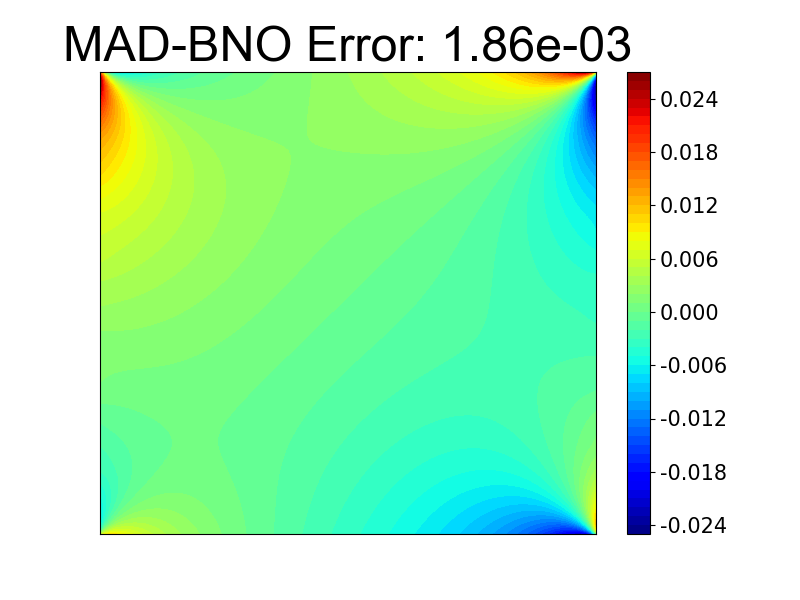}
		\caption{}
		\label{}
	\end{subfigure}
	\hfill
	\begin{subfigure}[b]{1\textwidth}
		\centering
		\includegraphics[width=\textwidth]{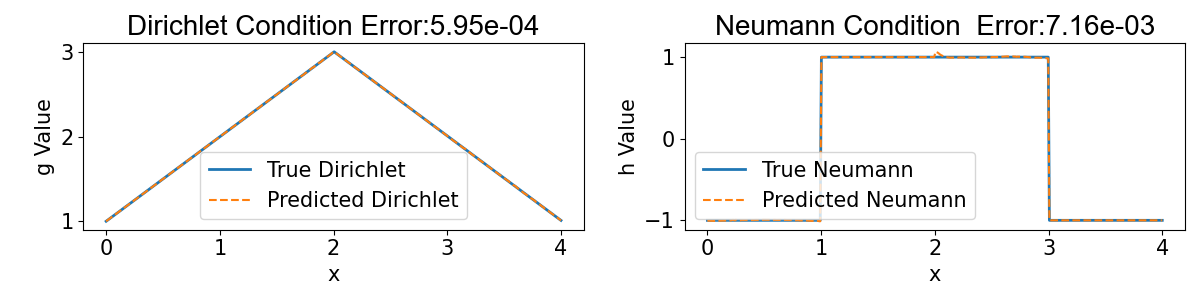}
		\caption{}
		\label{}
	\end{subfigure}
	\label{}
	\caption{Comparison of MAD-BNO and PI-DeepONet's predictions for the analytical solution  $\bm{u(x, y) = x+y+1} $ of the Laplace equation with $\partial \Omega_{D}=\Gamma_{1}\cup\Gamma_{3}$ in (Fig.~\ref{square4}). 
}\label{DN3}
\end{figure}

\begin{figure}[H]\label{DN figure3}
	\centering
	\begin{subfigure}[b]{0.3\textwidth}
		\centering
		\includegraphics[width=\textwidth]{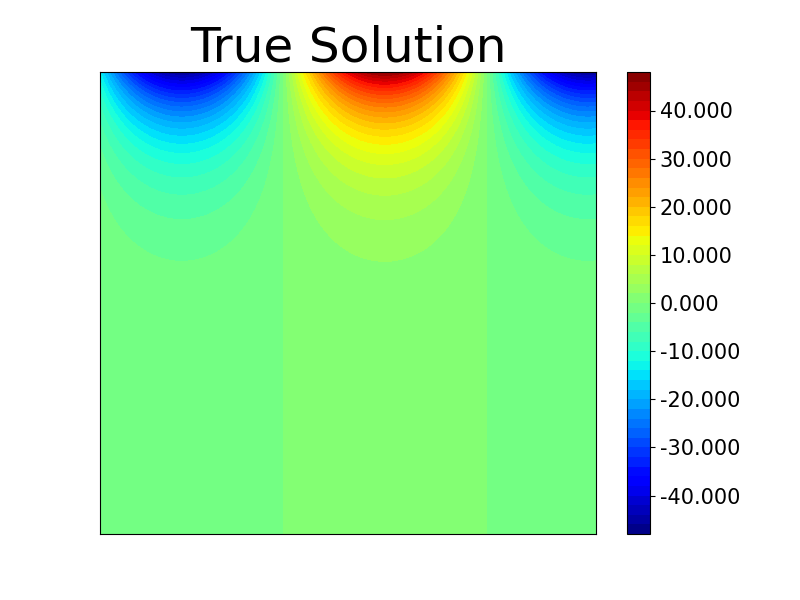}
		\caption{}
		\label{fig:image1}
	\end{subfigure}
	\hfill
	\begin{subfigure}[b]{0.3\textwidth}
		\centering
		\includegraphics[width=\textwidth]{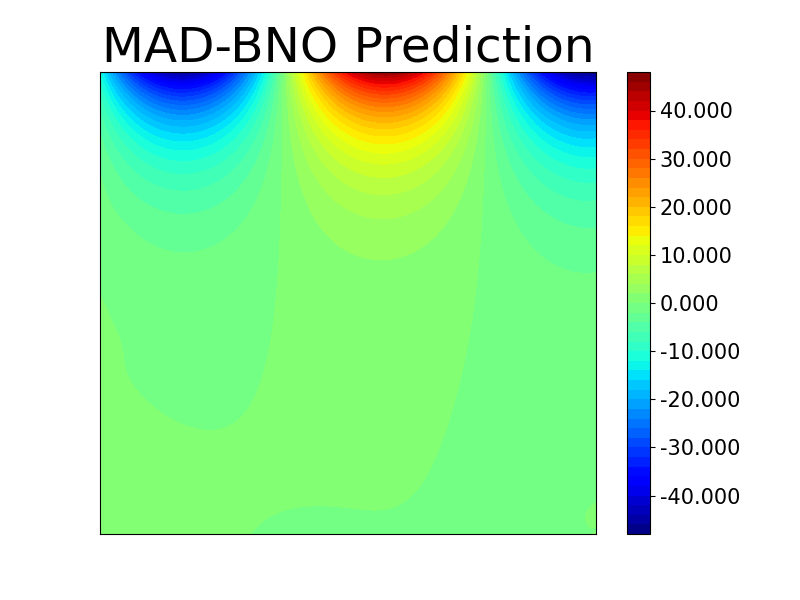}
		\caption{}
		\label{}
	\end{subfigure}
	\hfill
	\label{}
	\begin{subfigure}[b]{0.3\textwidth}
		\centering
		\includegraphics[width=\textwidth]{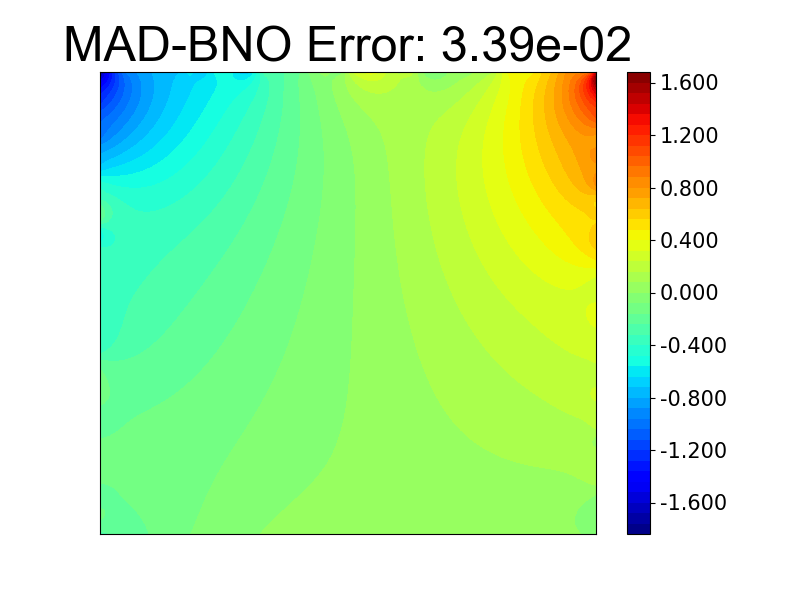}
		\caption{}
		\label{}
	\end{subfigure}
	\hfill
	\label{}
	\begin{subfigure}[b]{1\textwidth}
		\centering
		\includegraphics[width=\textwidth]{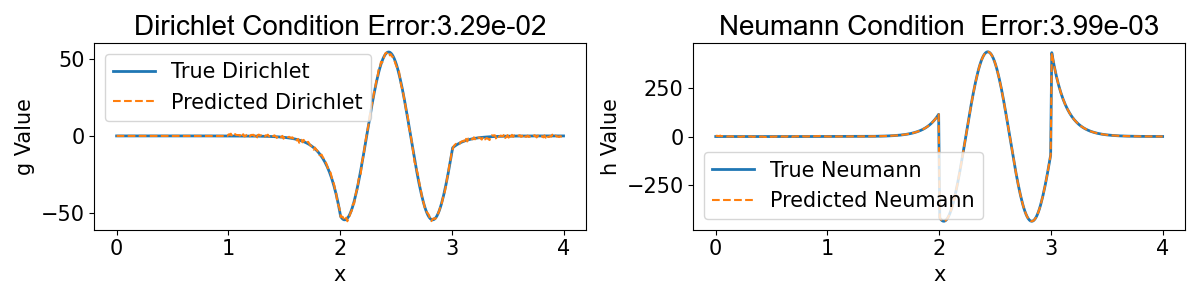}
		\caption{}
		\label{}
	\end{subfigure}
	\label{}
	\caption{Comparison of MAD-BNO and PI-DeepONet's predictions for the analytical solution   $\bm{u(x, y) = \sin(8x + 3)e^{8x - 4}}$  of the Laplace equation  with $\partial \Omega_{D}=\Gamma_{1}$ in (Fig.~\ref{square4}). 
}\label{DN4}
\end{figure}

  
﻿
Numerical results for the Laplace equation with mixed boundary conditions are summarized in Table~\ref{result2}, while detailed prediction outcomes for different boundary configurations are shown in Figs.~\ref{DN1}--\ref{DN4}, including the corresponding errors in the complementary boundary conditions. The results demonstrate that the proposed method consistently achieves an accuracy ranging from \(10^{-3}\) to \(10^{-2}\) in terms of the relative $L^2$ error of the reconstructed interior solution across all test cases.
﻿
\begin{table}[H]
	\centering
	\scriptsize 
	\begin{tabular}{c|c|c|c|c|c|c|c|c}
		\toprule
		Equation & Method & $ \partial \Omega_D$ & Training Time & Epoch & Training Loss & Dirichlet Error & Neumann Error & Total Error \\
		\midrule
		\multirow{2}{*}{$\Delta u=0$} 
		& MAD-BNO & $\Gamma_{1}$ & 1.62 h & 50000 & 7.49E-6 & 6.10E-3 & 1.35E-2 & 8.82E-3 \\
		& MAD-BNO & $\Gamma_{1}\cup\Gamma_{3}$ & 1.67 h & 50000 & 8.85E-6 & 7.23E-3 & 2.21E-2 & 9.13E-3\\
		\bottomrule
	\end{tabular}
	\caption{Performance of MAD-BNO for the Laplace equation (Eq.~\ref{eq:laplace}) under mixed boundary conditions, with different choices of Dirichlet boundary parts $\partial \Omega_D$ and Neumann boundary parts  $\partial \Omega_N$. 
		Reported errors are: (i) Dirichlet Error: relative error of $u$ over the entire boundary $\partial \Omega$ (including both given and predicted parts);  
		(ii) Neumann Error: relative error of $\tfrac{\partial u}{\partial n}$ over the entire boundary $\partial \Omega$ (including both given and predicted parts);  
		(iii) Total Error: relative  error of the reconstructed interior solution compared with the exact solution.  
		All reported errors denote mean values over the test set.
		}
	\label{result2}
\end{table}


  \subsection{Poisson equation}
  ~\par
\begin{figure}[H]
    \centering
    \begin{subfigure}[b]{0.3\textwidth}
        \centering
        \includegraphics[width=\textwidth]{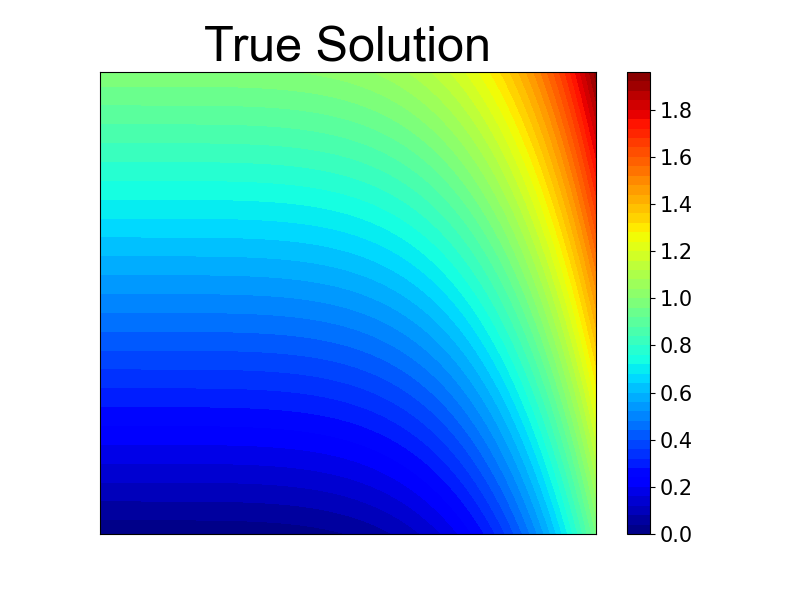}
        \caption{}
        \label{}
    \end{subfigure}
    \hfill
    \begin{subfigure}[b]{0.3\textwidth}
        \centering
        \includegraphics[width=\textwidth]{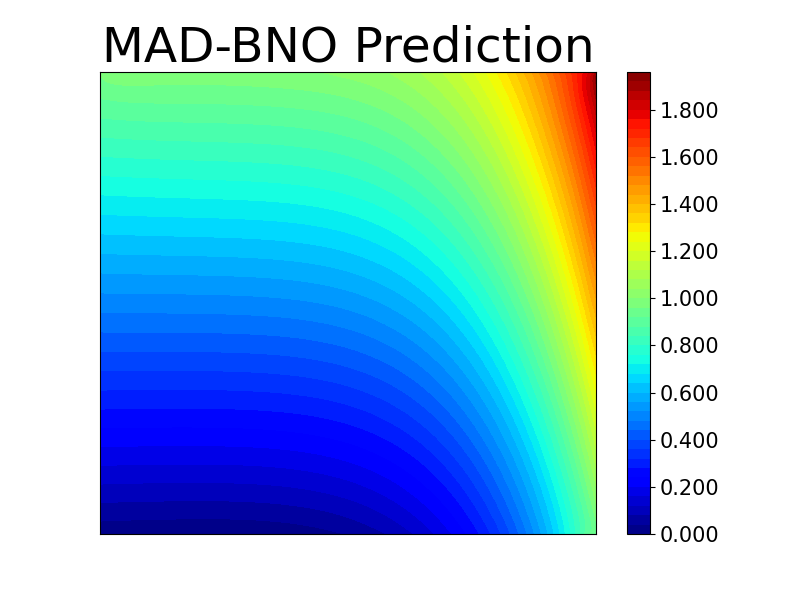}
        \caption{}
        \label{}
    \end{subfigure}
    \hfill
    \label{}
    \begin{subfigure}[b]{0.3\textwidth}
        \centering
        \includegraphics[width=\textwidth]{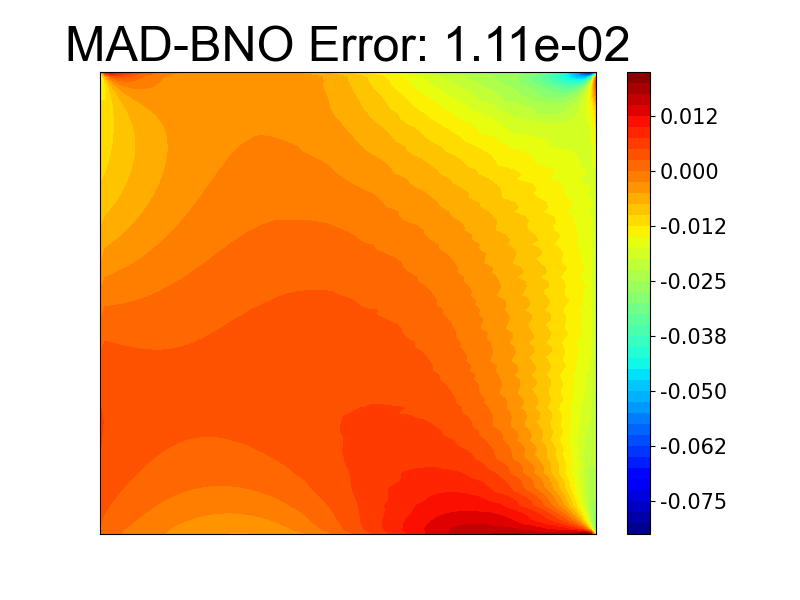}
        \caption{}
        \label{}
    \end{subfigure}
    \hfill
    \label{}
        \centering
    \begin{subfigure}[b]{0.3\textwidth}
        \centering
        \includegraphics[width=\textwidth]{poisson1_real.png}
        \caption{}
        \label{}
    \end{subfigure}
    \hfill
    \begin{subfigure}[b]{0.3\textwidth}
        \centering
        \includegraphics[width=\textwidth]{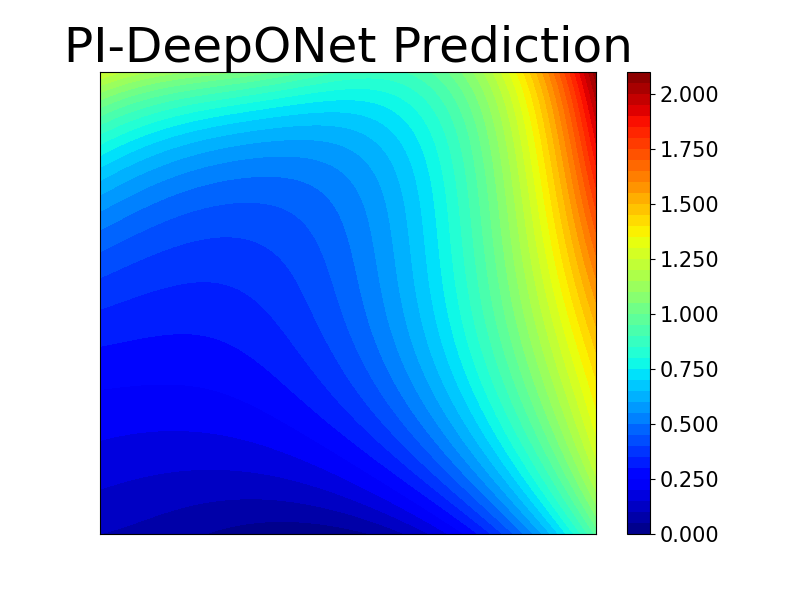}
        \caption{}
        \label{}
    \end{subfigure}
    \hfill
    \label{}
    \begin{subfigure}[b]{0.3\textwidth}
        \centering
        \includegraphics[width=\textwidth]{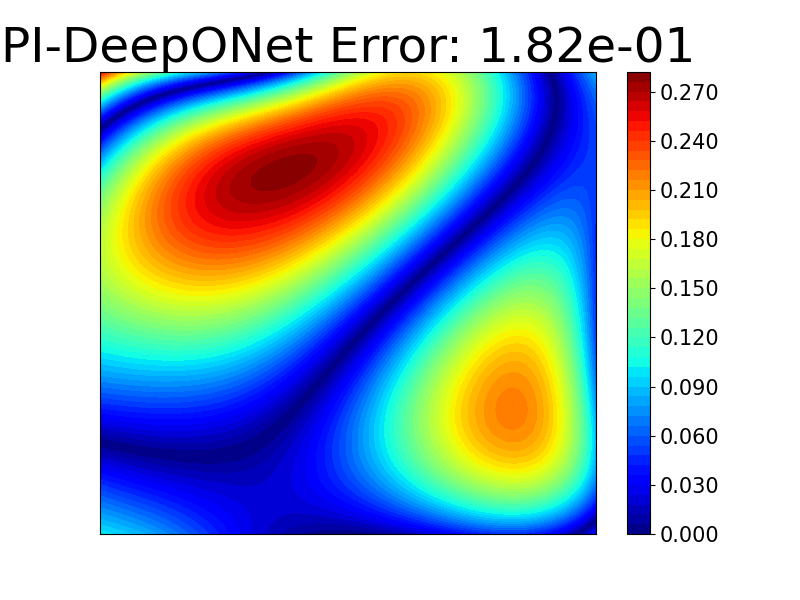}
        \caption{}
        \label{}
    \end{subfigure}
    \hfill
    \label{}
    \caption{Comparison of MAD-BNO and PI-DeepONet's predictions for the analytical solution   $\bm{ u(x, y) = x^{5}+y}$ of the Poisson equation(Eq.~\ref{eq:poisson}).
}\label{poisson1}
\end{figure}
\begin{figure}[H]
	\centering
	\begin{subfigure}[b]{0.3\textwidth}
		\centering
		\includegraphics[width=\textwidth]{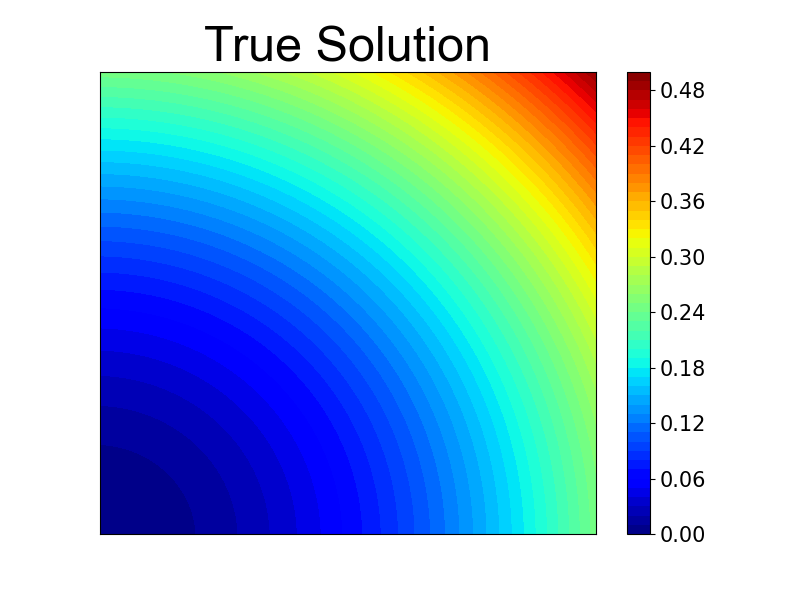}
		\caption{}
		\label{}
	\end{subfigure}
	\hfill
	\begin{subfigure}[b]{0.3\textwidth}
		\centering
		\includegraphics[width=\textwidth]{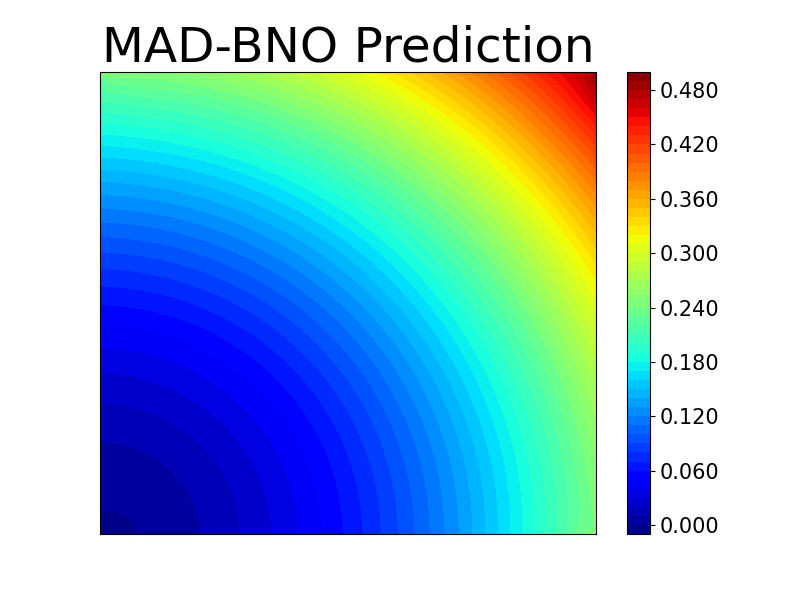}
		\caption{}
		\label{}
	\end{subfigure}
	\hfill
	\label{}
	\begin{subfigure}[b]{0.3\textwidth}
		\centering
		\includegraphics[width=\textwidth]{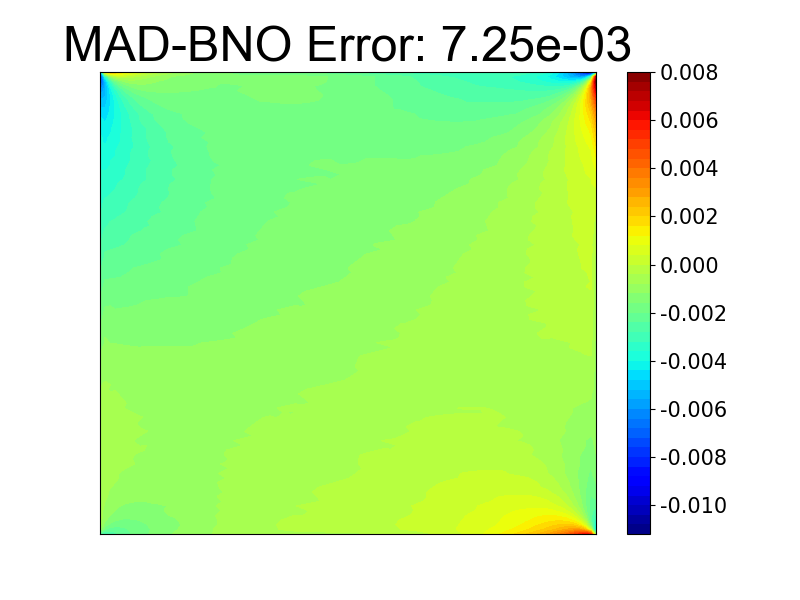}
		\caption{}
		\label{}
	\end{subfigure}
	\hfill
	\label{}
	\centering
	\begin{subfigure}[b]{0.3\textwidth}
		\centering
		\includegraphics[width=\textwidth]{poisson2_real.png}
		\caption{}
		\label{}
	\end{subfigure}
	\hfill
	\begin{subfigure}[b]{0.3\textwidth}
		\centering
		\includegraphics[width=\textwidth]{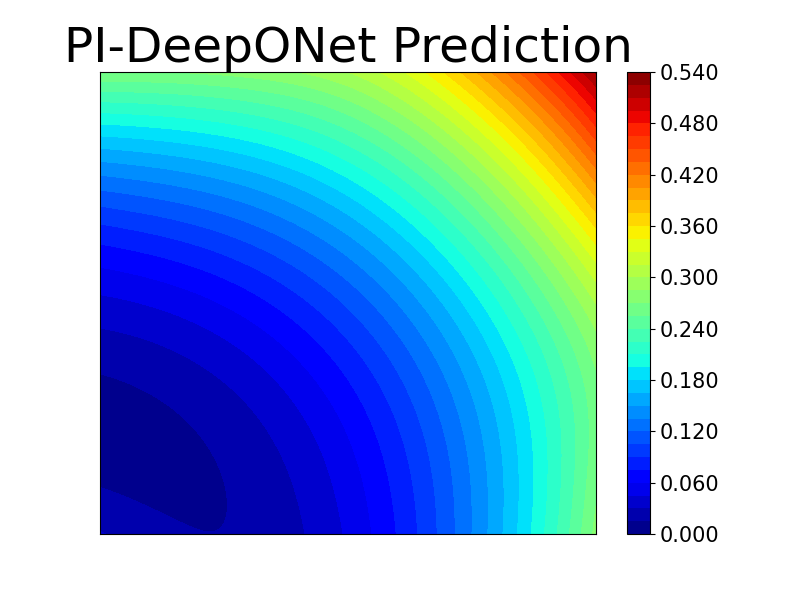}
		\caption{}
		\label{}
	\end{subfigure}
	\hfill
	\label{}
	\begin{subfigure}[b]{0.3\textwidth}
		\centering
		\includegraphics[width=\textwidth]{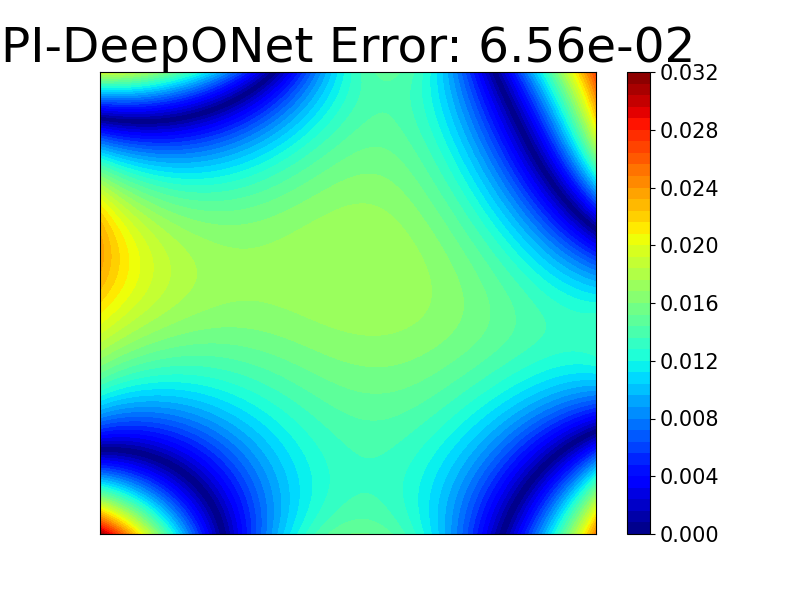}
		\caption{}
		\label{}
	\end{subfigure}
	\hfill
	\label{}
	\caption{Comparison of MAD-BNO and PI-DeepONet's predictions for the analytical solution   $\bm{u(x, y) = \frac{x^{2}+y^{2}}{4}}$ of the Poisson equation(Eq.~\ref{eq:poisson}).
	}\label{poisson2}
\end{figure}

In this subsection, we extend MAD-BNO to solve the Poisson equation with source terms under Dirichlet boundary conditions. As shown in the methodology, the Poisson problem (Eq. \ref{eq:poisson}) is reformulated as an equivalent Laplace problem (Eqs.~\ref{2}-\ref{eq:laplacetrans}) in addition to a deterministic volume integral part that is evaluated by an external numerical integration procedure. This reformulation highlights the flexibility of MAD-BNO in handling PDEs that can be decomposed into boundary-related and domain-related contributions, and also demonstrates the general applicability of the boundary-integral perspective beyond pure Laplace-type problems. In particular, the method preserves the boundary-driven learning structure while incorporating additional corrections through classical numerical techniques, which avoids overcomplicating the neural operator training. Figs. \ref{poisson1}–\ref{poisson2} illustrate the numerical prediction results for the Poisson equation, showing that MAD-BNO achieves consistent accuracy and stability, further supporting its robustness when applied to broader classes of linear elliptic PDEs. Specifically, we tested polynomial, linear, exponential, and trigonometric source terms, and the method demonstrated reliable performance across all cases. As shown in Table \ref{result1}, when trained for 50,000 epochs, MAD-BNO achieves lower training loss and improved test performance compared with PI-DeepONet, while the training time is reduced by 1–2 orders of magnitude, indicating that the network has sufficiently converged. This training setup provides a fair and reproducible basis for comparing MAD-BNO with alternative methods.

\subsection{Helmholtz equation}
\subsubsection{Two-dimensional case}
Similar to the Laplace equation, the solution to the source-free Helmholtz equation can be represented and computed via boundary integral formulations. The main difference lies in the fundamental solution: while the Laplace equation uses a logarithmic kernel, the Helmholtz equation employs a complex-valued Hankel function that naturally captures the oscillatory behavior of waves. We learn the neural operator for the Helmholtz equation with various wavenumbers, specifically \(k=1\), \(k=10\), and \(k=100\) in Eq.~\ref{eq:helmholtz}, to assess performance at different frequencies. To further evaluate robustness and generalization, we tested multiple classes of boundary inputs, including trigonometric products and combinations of trigonometric and exponential functions. The networks were trained for 50,000 epochs, which ensured sufficient convergence of the training loss; as shown in Table \ref{result1}, MAD-BNO consistently outperforms PI-DeepONet across all test cases, while the required training time is reduced by roughly one order of magnitude.
﻿
\begin{figure}[H]
	\centering
	\begin{subfigure}[b]{0.3\textwidth}
		\centering
		\includegraphics[width=\textwidth]{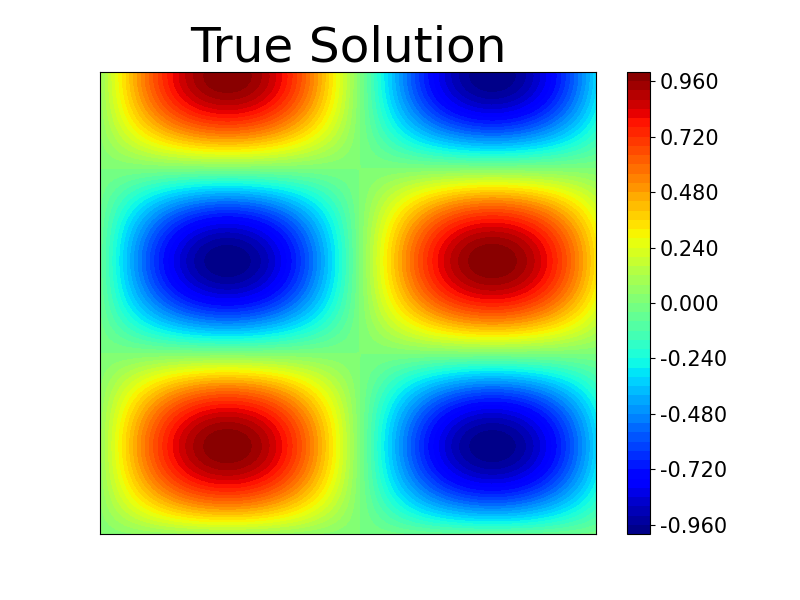}
		\caption{}
		\label{}
	\end{subfigure}
	\hfill
	\begin{subfigure}[b]{0.3\textwidth}
		\centering
		\includegraphics[width=\textwidth]{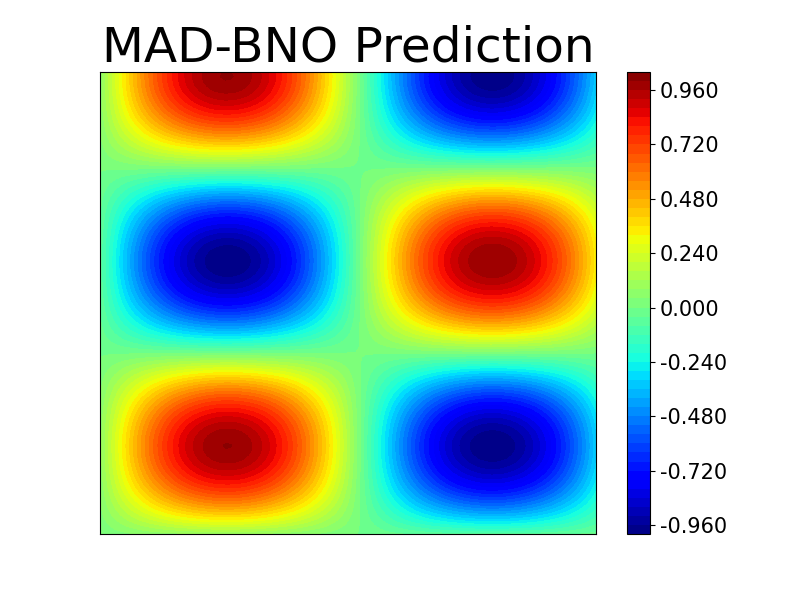}
		\caption{}
		\label{}
	\end{subfigure}
	\hfill
	\label{}
	\begin{subfigure}[b]{0.3\textwidth}
		\centering
		\includegraphics[width=\textwidth]{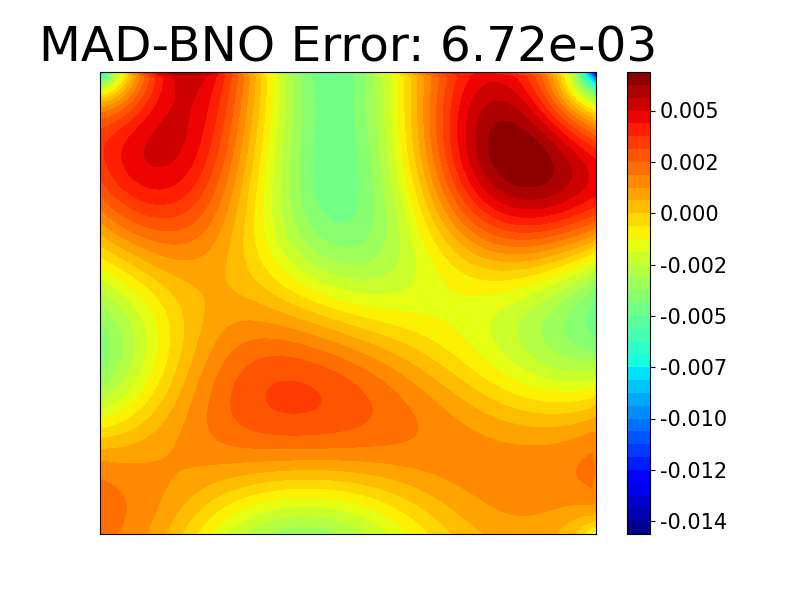}
		\caption{}
		\label{}
	\end{subfigure}
	\hfill
	\label{}
		\centering
	\begin{subfigure}[b]{0.3\textwidth}
		\centering
		\includegraphics[width=\textwidth]{H-100-1-R.png}
		\caption{}
		\label{}
	\end{subfigure}
	\hfill
	\begin{subfigure}[b]{0.3\textwidth}
		\centering
		\includegraphics[width=\textwidth]{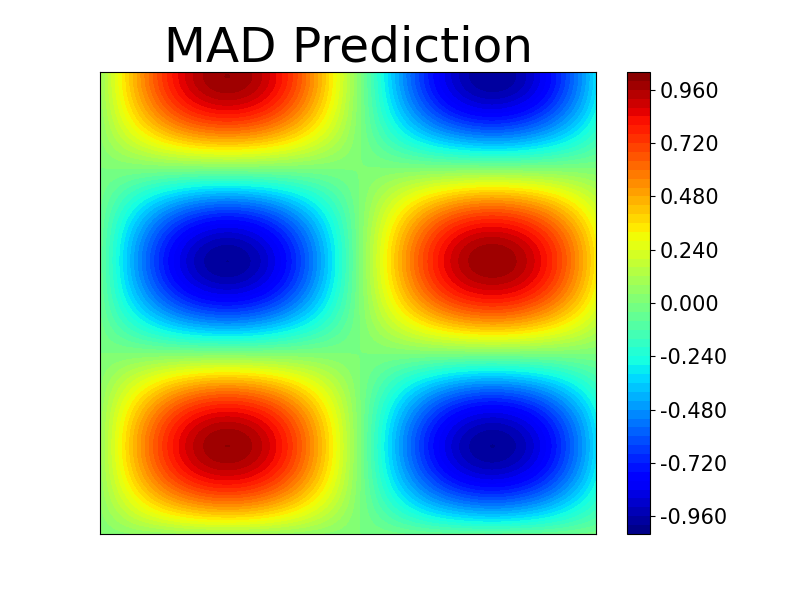}
		\caption{}
		\label{}
	\end{subfigure}
	\hfill
	\label{}
	\begin{subfigure}[b]{0.3\textwidth}
		\centering
		\includegraphics[width=\textwidth]{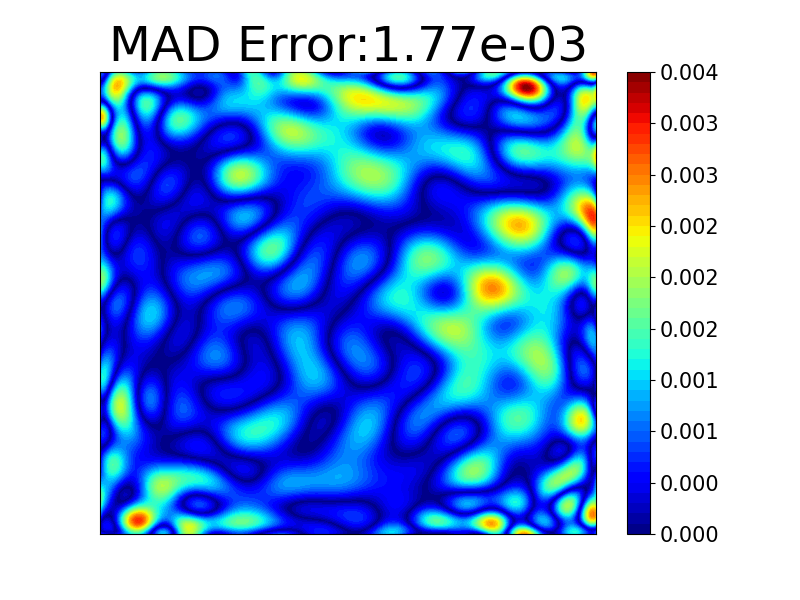}
		\caption{}
		\label{}
	\end{subfigure}
	\hfill
	\label{}
		\centering
	\begin{subfigure}[b]{0.3\textwidth}
		\centering
		\includegraphics[width=\textwidth]{H-100-1-R.png}
		\caption{}
		\label{}
	\end{subfigure}
	\hfill
	\begin{subfigure}[b]{0.3\textwidth}
		\centering
		\includegraphics[width=\textwidth]{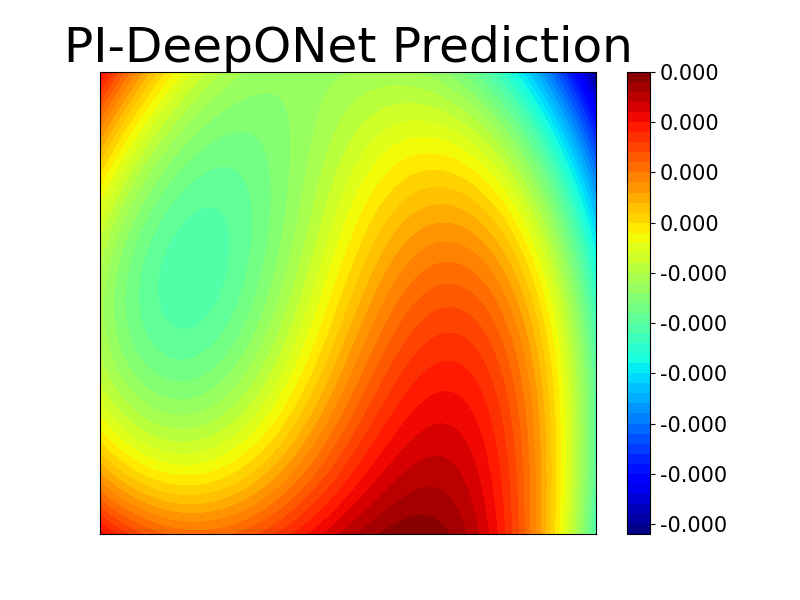}
		\caption{}
		\label{}
	\end{subfigure}
	\hfill
	\label{}
	\begin{subfigure}[b]{0.3\textwidth}
		\centering
		\includegraphics[width=\textwidth]{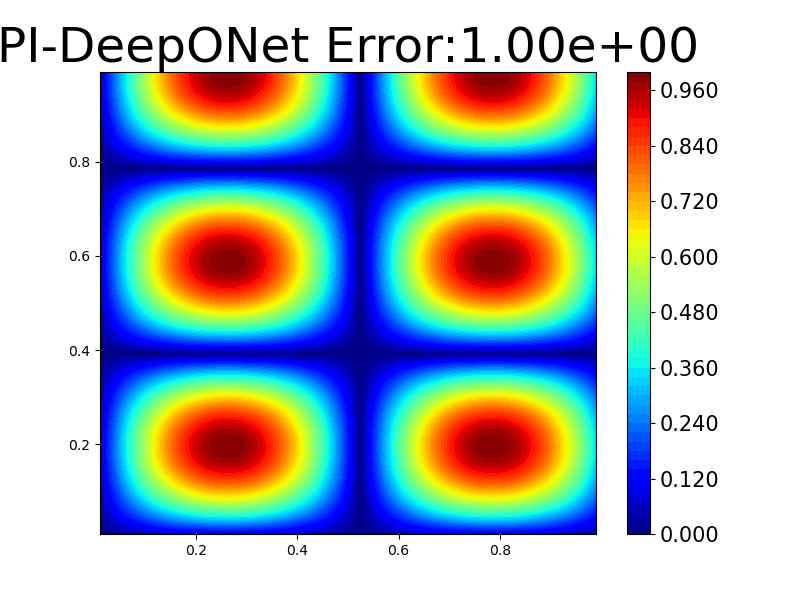}
		\caption{}
		\label{}
	\end{subfigure}
	\hfill
	\label{}
	\caption{Comparison of MAD-BNO, MAD, and PI-DeepONet's predictions for the analytical solution $\bm{u(x, y) = \sin(6x)\sin(8y)}$ of the Helmholtz equation (Eq.~\ref{eq:helmholtz}) with $k=10$ in the square domain $(0,1)\times(0,1)$.
}\label{h1}
\end{figure}
\begin{figure}[H]
	\centering
	\begin{subfigure}[b]{0.3\textwidth}
		\centering
		\includegraphics[width=\textwidth]{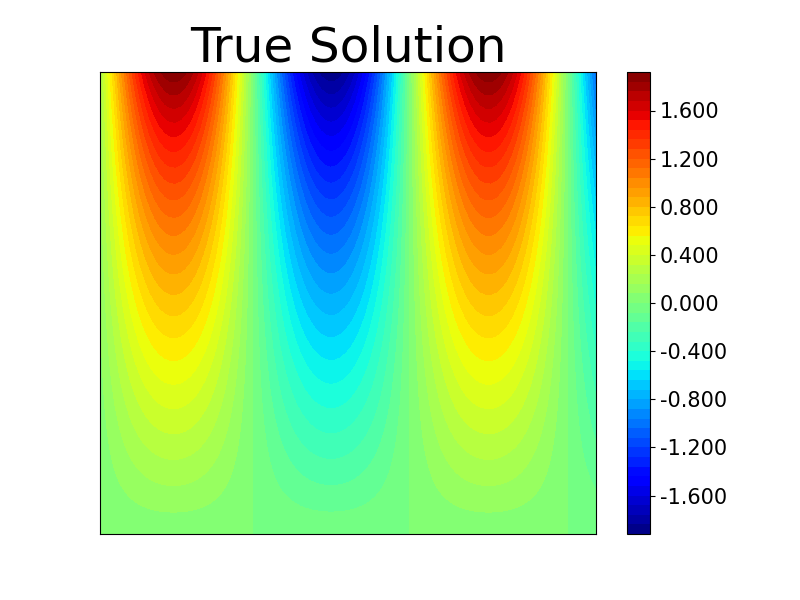}
		\caption{}
		\label{}
	\end{subfigure}
	\hfill
	\begin{subfigure}[b]{0.3\textwidth}
		\centering
		\includegraphics[width=\textwidth]{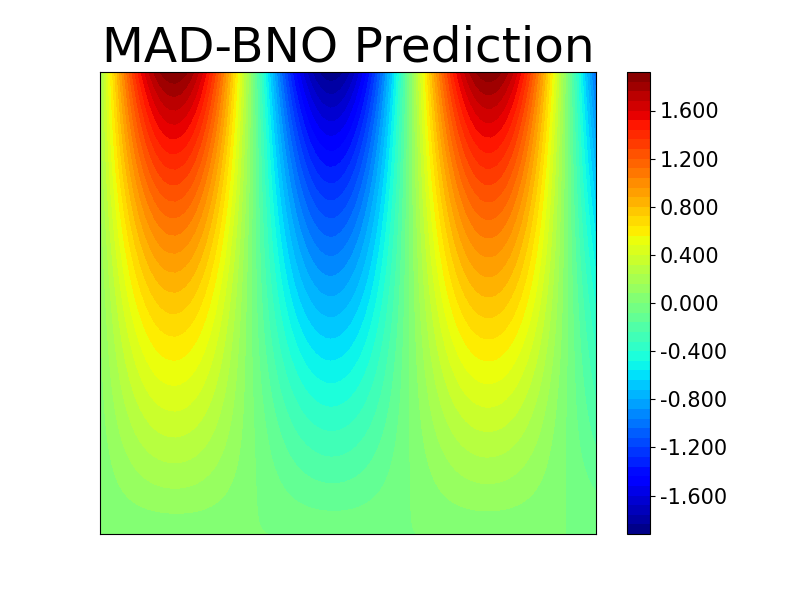}
		\caption{}
		\label{}
	\end{subfigure}
	\hfill
	\label{}
	\begin{subfigure}[b]{0.3\textwidth}
		\centering
		\includegraphics[width=\textwidth]{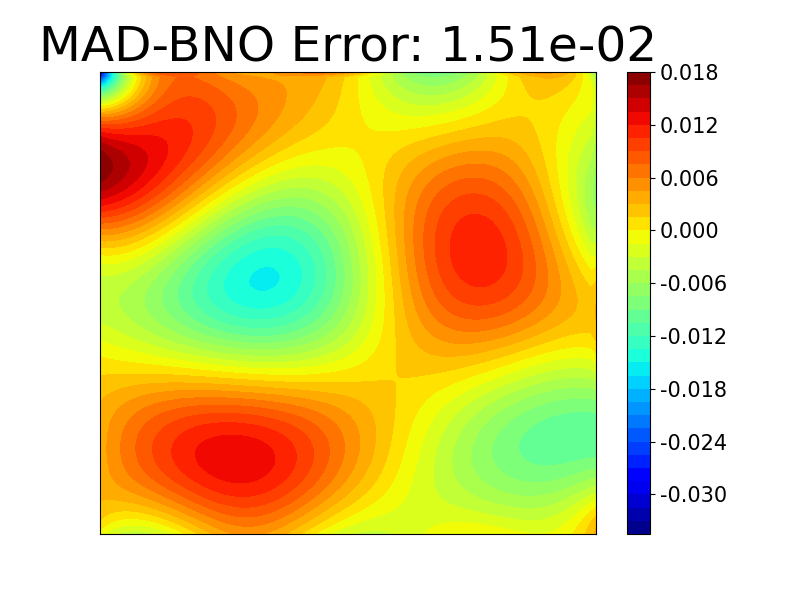}
		\caption{}
		\label{}
	\end{subfigure}
	\hfill
	\label{}
	\centering
	\begin{subfigure}[b]{0.3\textwidth}
		\centering
		\includegraphics[width=\textwidth]{H-100-3-R.png}
		\caption{}
		\label{}
	\end{subfigure}
	\hfill
	\begin{subfigure}[b]{0.3\textwidth}
		\centering
		\includegraphics[width=\textwidth]{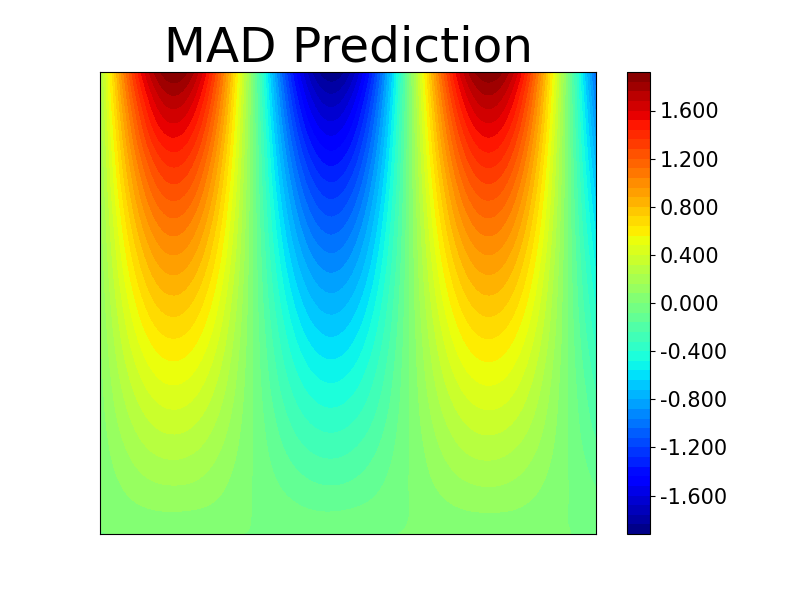}
		\caption{}
		\label{}
	\end{subfigure}
	\hfill
	\label{}
	\begin{subfigure}[b]{0.3\textwidth}
		\centering
		\includegraphics[width=\textwidth]{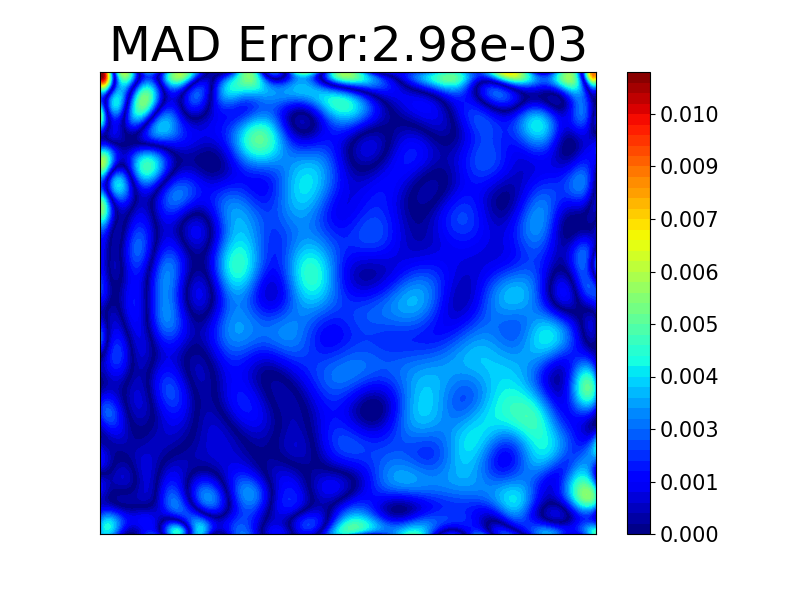}
		\caption{}
		\label{}
	\end{subfigure}
	\hfill
	\label{}
	\centering
	\begin{subfigure}[b]{0.3\textwidth}
		\centering
		\includegraphics[width=\textwidth]{H-100-3-R.png}
		\caption{}
		\label{}
	\end{subfigure}
	\hfill
	\begin{subfigure}[b]{0.3\textwidth}
		\centering
		\includegraphics[width=\textwidth]{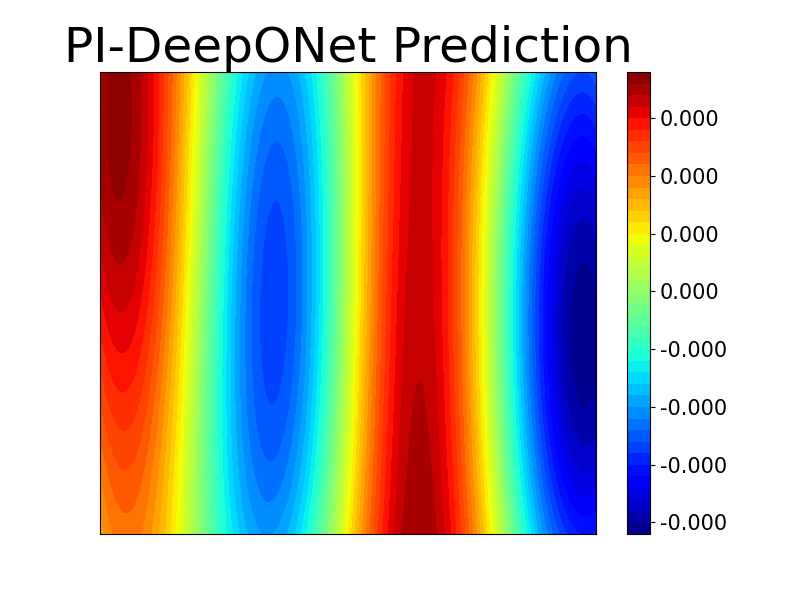}
		\caption{}
		\label{}
	\end{subfigure}
	\hfill
	\label{}
	\begin{subfigure}[b]{0.3\textwidth}
		\centering
		\includegraphics[width=\textwidth]{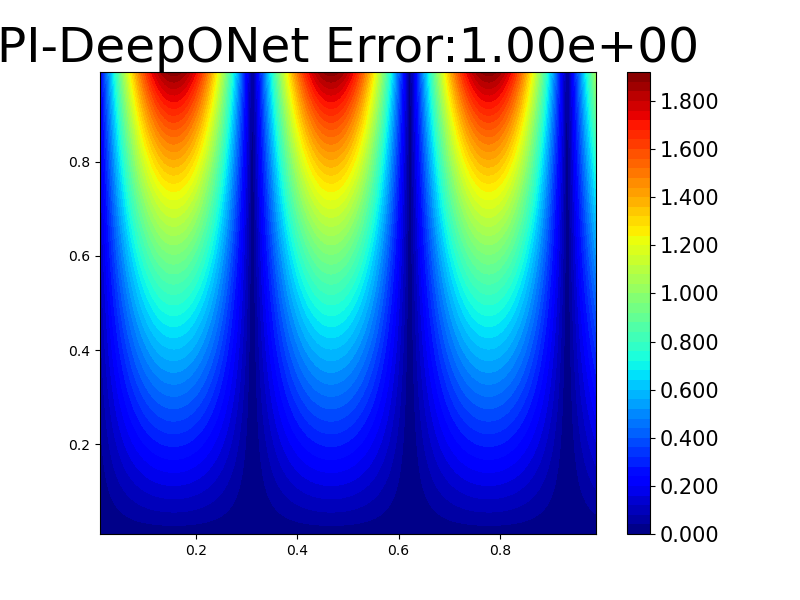}
		\caption{}
		\label{}
	\end{subfigure}
	\hfill
	\label{}
	\caption{Comparison of MAD-BNO, MAD, and PI-DeepONet's predictions for the analytical solution   $\bm{ u(x, y) =sin(\sqrt{102}x)sinh(\sqrt{2}y)}$ of the Helmholtz equation(Eq.~\ref{eq:helmholtz}) with $k=10$.
}\label{h100}
\end{figure}
\begin{figure}[H]
	\centering
	\begin{subfigure}[b]{0.3\textwidth}
		\centering
		\includegraphics[width=\textwidth]{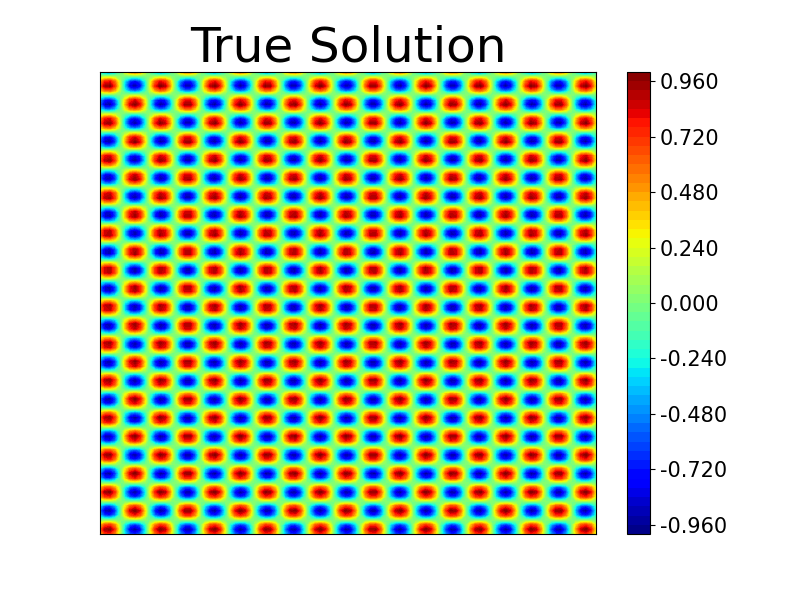}
		\caption{}
		\label{}
	\end{subfigure}
	\hfill
	\begin{subfigure}[b]{0.3\textwidth}
		\centering
		\includegraphics[width=\textwidth]{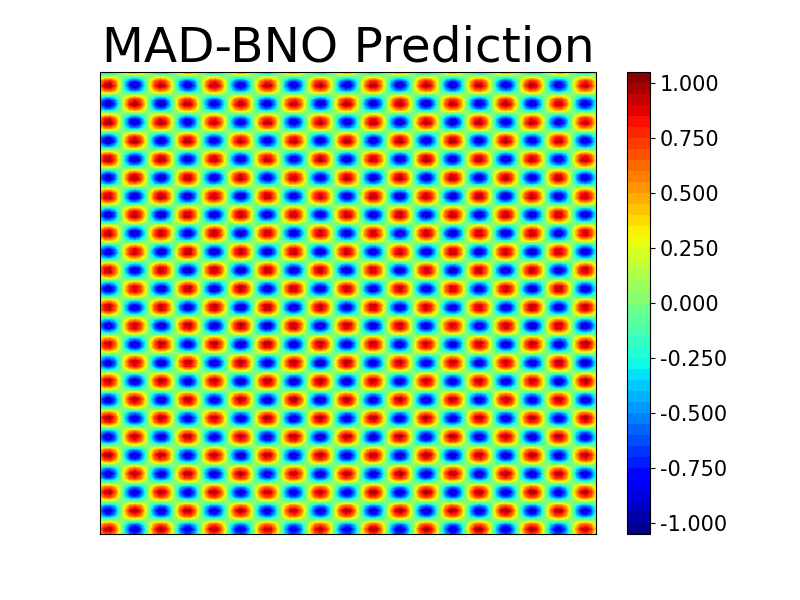}
		\caption{}
		\label{}
	\end{subfigure}
	\hfill
	\label{}
	\begin{subfigure}[b]{0.3\textwidth}
		\centering
		\includegraphics[width=\textwidth]{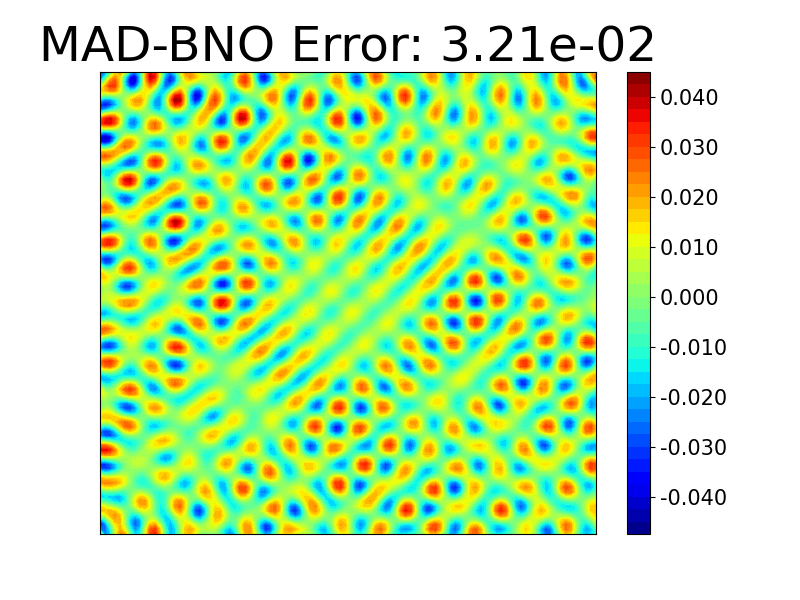}
		\caption{}
		\label{}
	\end{subfigure}
	\hfill
	\label{}
		\begin{subfigure}[b]{1\textwidth}
		\centering
		\includegraphics[width=\textwidth]{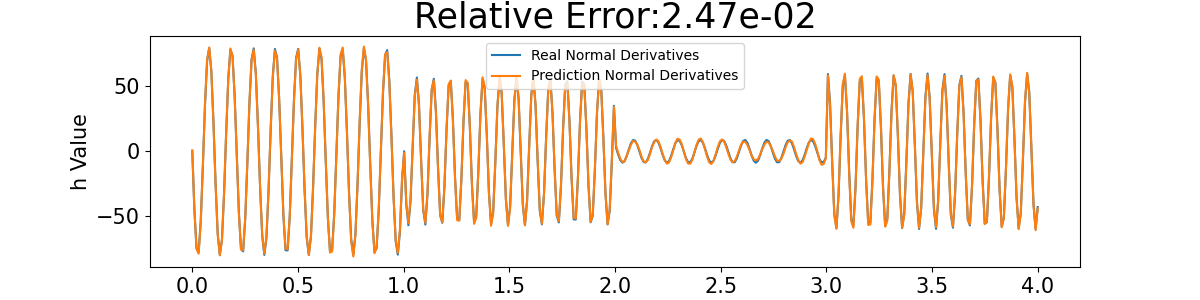}
		\caption{}
		\label{}
	\end{subfigure}
	\hfill
	\label{}
	\caption{MAD-BNO prediction for the analytical solution $\bm{ u(x, y) = \sin(60x)\sin(80y) }$ of the Helmholtz equation (Eq.~\ref{eq:helmholtz}) with $k=100$. \textbf{(d)}:Comparison of the Neumann boundary values predicted by MAD-BNO with the exact values, where the square boundary of the domain is parameterized into the interval $[0,4]$ by traversing counterclockwise from the point $(0,0)$.}\label{H10000}
\end{figure}
\subsubsection{Three-dimensional case}
To illustrate the extensibility of the proposed method to three-dimensional problems, we consider the Helmholtz equation with Dirichlet boundary conditions in a 3D domain. In this setting, the fundamental solution of the three-dimensional Helmholtz operator $(\Delta + k^2)$ is given by  
\begin{equation}
	G(\mathbf{r}) = \frac{e^{ik|\mathbf{r}|}}{4\pi |\mathbf{r}|},
\end{equation}
which acts as the kernel function in the boundary integral formulation. The computational domain is discretized by sampling 1200 points on the surface, which serve as the boundary collocation points in the boundary integral formulation. Numerical results indicate that the accuracy of normal derivative predictions in three dimensions is comparable to that achieved in two dimensions. \par
\begin{figure}[H]
	\centering
	\includegraphics[width=0.5\linewidth]{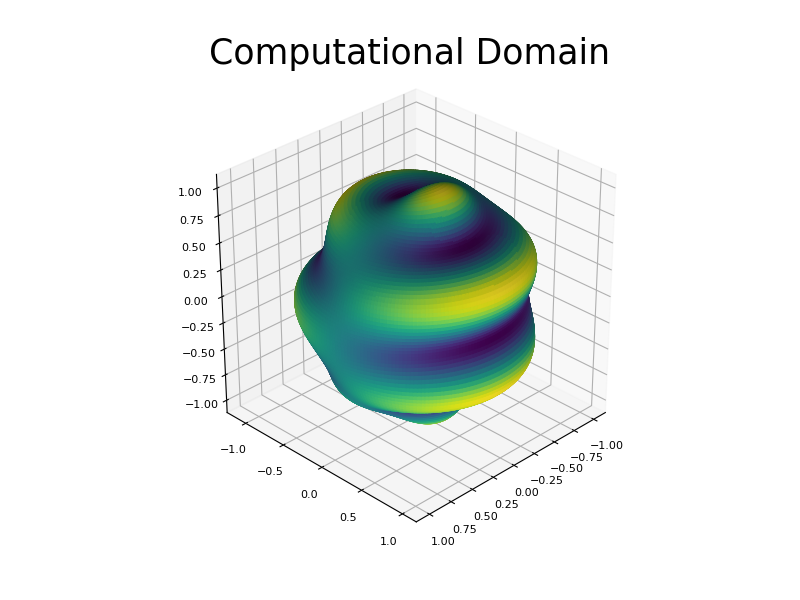}
	\caption{Computational Domain for the 3D Helmholtz Equation}
	\label{3D}
\end{figure}
\begin{figure}[H]
	\centering
	\begin{subfigure}[b]{0.3\textwidth}
		\centering
		\includegraphics[width=\textwidth]{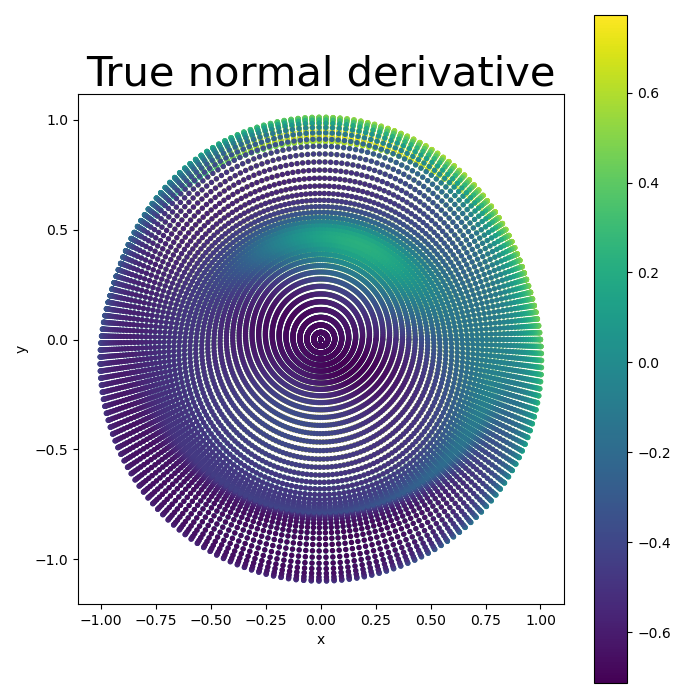}
		\caption{}
		\label{}
	\end{subfigure}
	\hfill
	\begin{subfigure}[b]{0.3\textwidth}
		\centering
		\includegraphics[width=\textwidth]{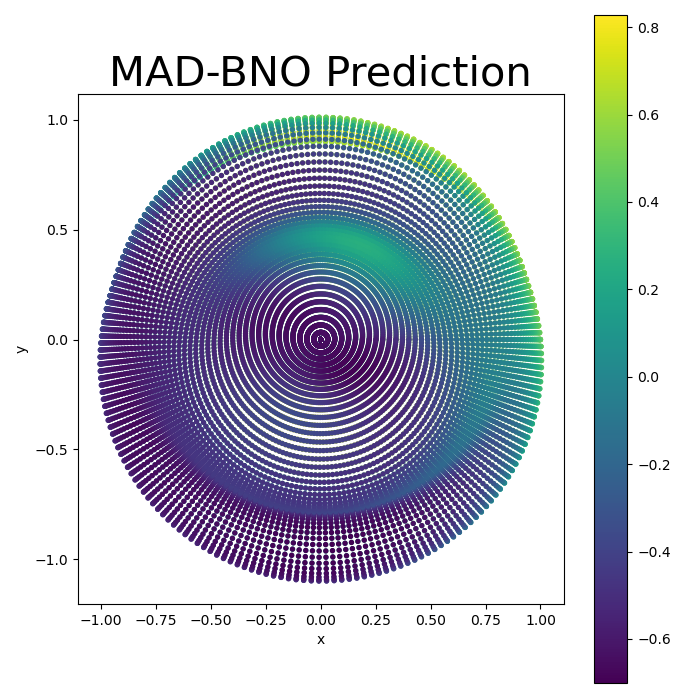}
		\caption{}
		\label{}
	\end{subfigure}
	\hfill
	\label{}
	\begin{subfigure}[b]{0.3\textwidth}
		\centering
		\includegraphics[width=\textwidth]{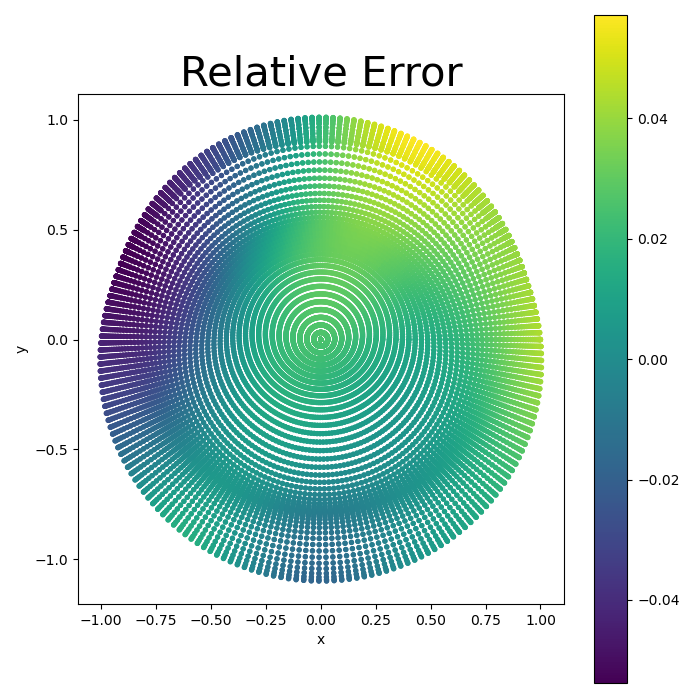}
		\caption{}
		\label{}
	\end{subfigure}
	\hfill
	\label{}
	\caption{MAD-BNO prediction for the analytical solution $sin(\sqrt{0.2}x+\sqrt{0.3}y+\sqrt{0.5}z)$ of the Helmholtz equation (Eq.~\ref{eq:helmholtz}) with $k=1$. }\label{3Dnorm}
\end{figure}

Figure~\ref{3D} illustrates the three-dimensional computational domain used in our study. Figure \ref{3Dnorm} shows the top-down view of the normal derivative, the predicted values, and the corresponding relative $L^{2}$ error. Table~\ref{result3} reports the performance of MAD-BNO in the 3D setting, showing that the accuracy remains at the same order as in the two-dimensional case. In this work, we do not perform three-dimensional integration; instead, we only compute the $L^2$ errors on the boundary. For future practical applications in 3D, the method can be extended using boundary integral formulations, and the evaluation of both boundary and domain integrals can be accelerated with fast algorithms such as the fast multipole method (FMM)
\cite{FMM1,FMM2}.

	\begin{table}[H]
		\centering
		\small
		\setlength{\tabcolsep}{6pt} % 调整列间距
		\renewcommand{\arraystretch}{0.9} % 增加行高
		\begin{tabular}{c|c|c|c|c|c}
			\toprule
			Equation& Method & Training Time  &Epoch& Training Loss&$\frac{\partial u}{\partial n}$ Error \\
			\midrule
		              
		{$\Delta u+u=0$} & MAD-BNO &22.37h &50000& 9.88e-6 &1.68e-2 \\
			\bottomrule
		\end{tabular}
		\caption{Performance  of MAD-BNO in operator learning for the three-dimensional Helmholtz Equations with Dirichlet boundary conditions.
		}\label{result3}
	\end{table}
This example demonstrates the applicability of the proposed method to three-dimensional problems, while more complex scenarios in 3D are reserved for future investigation.
\section{Discussion}
\label{sec:discussion}

﻿
By employing our proposed artificial data generation method and leveraging boundary integral representations, we transform PDE operator learning for linear elliptic equations with fundamental solutions into a physics-embedded, boundary-centric data-driven training framework. This framework obviates the need for both expensive external data and full-domain interior sampling, thereby achieving orders-of-magnitude higher efficiency in operator learning while maintaining comparable or superior accuracy due to the high-fidelity synthetic data generated through boundary-consistent physics constraints.  Focusing collocation on surfaces rather than volumes markedly
reduces the number of training points and mitigates the combinatorial growth of sampling cost with dimension and mesh resolution. The formulation is also broadly applicable within the same mathematical regime. Boundary collocation extends directly to three dimensions and to other linear problems with fundamental solutions---for example, the linear Poisson-Boltzmann equation, Stokes equation and Helmholtz equation. Complex geometries are naturally accommodated through surface discretizations, further lowering computational demands without redesigning the learning pipeline.\par

The efficiency of MAD-BNO can also be interpreted in terms of linear algebra. 
Traditional PDE solvers typically discretize the PDE to obtain a linear system 
\[A \mathbf{u} = \mathbf{b},\] where $A$ is the system matrix arising from finite difference, finite element, or spectral discretization, and $\mathbf{b}$ encodes boundary conditions and source terms. Classical iterative solvers must recompute $\mathbf{u}$ for each new $\mathbf{b}$, which is computationally expensive. In contrast, the trained neural operator learns the mapping from boundary conditions to the solution, which is mathematically analogous to applying $A^{-1}$, though not computed explicitly. More generally, all neural operator learning methods aim to find a neural network that approximates the inverse operator $\mathcal{L}^{-1}$ of a given operator $\mathcal{L}$. Once trained, MAD-BNO can generate solutions for new inputs without solving the linear system again, enabling repeated inference efficiently.\par

These choices translate directly into practical efficiency. The interior field can be reconstructed from boundary and (if source exists) volume integrals at inference time. However, for large-scale problems, direct evaluation of boundary  and volume integrals still scales as $O(N^2)$ and may become costly. By further employing fast algorithms---such as the Fast Multipole Method (FMM) \cite{FMM1,FMM2}---to accelerate boundary and volume integral evaluations, the complexity can be reduced to nearly $O(N)$. This makes MAD-BNO inherently scalable to real-time and large-scale applications.\par
﻿
Finally, MAD-BNO relies only on standard components---boundary values and mean squared error (MSE) losses---which makes it conceptually simple and lightweight. While most existing neural operator methods are voxel- or grid-based, our boundary-centric formulation offers a complementary perspective. Potential synergies with modern data-driven techniques (e.g., operator priors, active sampling, advanced network architectures) remain to be investigated. Future work will further explore extensions of this boundary-centric paradigm, with the aim of broadening its applicability across diverse PDEs and computational settings.

%\appendix
%\section{Example Appendix Section}
%\label{app1}
%Supplementary information is provided as a separate file
%and is available for this paper.

\section*{CRediT authorship contribution statement}
\textbf{B.Z.L.} conceived and designed the study and supervised
the project. \textbf{H.C.W.} implemented the numerical methods and conducted the experiments. \textbf{H.W.} contributed by providing comparative data for validating the proposed method. All authors contributed to analyzing the data and revising the manuscript.

\section*{Data availability}
		Code and data will be made available on request.
\section*{Declaration of competing interest}
The authors declare no competing interests.
\section*{Acknowledgements}
This research was funded by the Strategic Priority
Research Program of the Chinese Academy of Sciences (Grant No.XDB0500000) and the National Natural Science Foundation of China (Grant No.12371413).\par{\tiny }
The numerical calculations in this study were carried
out on the ORISE Supercomputer.

\bibliographystyle{unsrt}
\bibliography{reference}
\end{document}